\documentclass{article}

\usepackage[utf8]{inputenc}
\usepackage{CJKutf8}

    \PassOptionsToPackage{numbers, sort&compress}{natbib}

 \usepackage[main, final]{neurips_2026}

\usepackage[utf8]{inputenc} 
\usepackage[T1]{fontenc}    
\usepackage{hyperref}       
\usepackage{url}            
\usepackage{booktabs}       
\usepackage{amsfonts}       
\usepackage{nicefrac}       
\usepackage{microtype}      
\usepackage{xcolor}         

\usepackage{graphicx}
\usepackage{amsmath}
\usepackage{subcaption}
\usepackage{amssymb}
\usepackage{mathtools}
\usepackage{amsthm}
\usepackage{multirow}
\usepackage{enumitem}
\usepackage{wrapfig}
\usepackage{tipa}

\usepackage[capitalize,noabbrev, nameinlink]{cleveref}
\makeatletter
\AddToHook{cmd/appendix/before}{\def\cref@section@alias{appendix}\def\cref@subsection@alias{appendix}}
\makeatother

\usepackage{algorithm}
\usepackage{algorithmic}
\usepackage{array}
\usepackage[most]{tcolorbox}

\newcolumntype{H}{>{\setbox0=\hbox\bgroup}c<{\egroup}@{}}
\newtcolorbox{promptbox}[1][]{
    colback=gray!5!white,
    colframe=gray!75!black,
    fonttitle=\bfseries,
    title=#1,
    arc=2pt,
    outer arc=2pt,
    left=5pt,
    right=5pt,
    top=5pt,
    bottom=5pt,
    boxrule=0.5pt
}

\title{Fine-Tuning Language Models to Know \\ What They Know}

\author{
Sangjun Park\textsuperscript{1,2}\quad
Elliot Meyerson\textsuperscript{2}\quad
Xin Qiu\textsuperscript{2}\quad 
Risto Miikkulainen\textsuperscript{1,2}\quad
\vspace{1mm} \\
\textsuperscript{1}The University of Texas at Austin\qquad 
\textsuperscript{2}Cognizant AI Lab
}

\begin{document}

\maketitle

\begin{abstract}
  Evaluating true metacognition in Large Language Models (LLMs) is difficult due to biases and heuristics. This paper presents a framework to measure and enhance LLM metacognition while controlling for these biases. A measurement method using the $d'_{\rm type2}$ metric is established to isolate metacognitive ability. The Evolution Strategy for Metacognitive Alignment (ESMA) is proposed, demonstrating robust generalization across unseen datasets, languages, and newly acquired knowledge. Gradual parameter updates reveal that these improvements are driven by a sparse set of parameters, opening a potential avenue for targeted metacognitive optimization.

\end{abstract}

\section{Introduction}
\label{sec:introduction}
Humans have extensive metacognitive abilities, which enable them to monitor and regulate their own cognitive processes \cite{Brown1987MetacognitionEC, chi1989, moshman2004}. A central component is self-knowledge, the capacity to assess what one knows \citep{Frith2012-ia, Binder2011-qr}. However, measuring self-knowledge is inherently challenging because observed behavior reflects not only access to internal states but also response biases and heuristics. For example, people may report low confidence for questions that appear difficult, regardless of their actual knowledge. While such heuristics can improve apparent metacognitive performance, they do not necessarily indicate genuine introspective access.
Psychological research has therefore devoted considerable effort to disentangling genuine metacognitive sensitivity from these confounds \citep{MANISCALCO2012422, Bjork2012-jt}.

This challenge extends directly to Large Language Models (LLMs) \cite{gemini25, qwen3}. LLM metacognition is often discussed in relation to practical goals such as calibrating confidence and reducing hallucinations. In these settings, using heuristics such as task difficulty can make responses more reliable. However, such improvements do not imply that a model is accessing its own internal knowledge state. To study metacognition itself, it is necessary to disentangle self-knowledge from heuristic-driven responses. 

To address this challenge, this work introduces a framework to evaluate LLM metacognition by comparing factual accuracy with self-reports elicited in independent contexts. Evolution Strategy for Metacognitive Alignment (ESMA; \textipa{/"EsmA:/})\footnote{For reproducibility, the code and experimental scripts are provided at: \url{https://github.com/cosmoquester/ESMA}.} is proposed as a fine-tuning method to improve metacognitive ability. To rigorously demonstrate the effectiveness of ESMA, this study systematically controls for the confounding factors that can artificially inflate apparent metacognitive performance.

First, to counter response biases, the $d'_{\rm type2}$ metric is employed within a dual-questioning protocol, isolating metacognitive functions from biased response tendencies.
Type 2 area under the receiver operating characteristic curve (AUROC) analysis consistently supports the results, ensuring that improvements in $d'_{\rm type2}$ reflect stronger metacognitive discrimination across continuous confidence.

Second, an \textit{I don't know} (IDK) unified prompt experiment uses the IDK alignment metric to test for prompt-template bias in a format never used during ESMA training. Moreover, the all alignment metric mitigates the illusion of knowing bias by requiring agreement between direct answer correctness and self-reports across the dual-question and IDK formats, following psychological methodologies.

Third, to rule out the cue-familiarity heuristic, the model learns fictional facts through supervised fine-tuning on FictionalQA before ESMA training on TriviaQA. The results show improved metacognitive monitoring of newly acquired knowledge without prior exposure to the corresponding meta questions.

Finally, model-specific confounds, such as benchmark-specific shortcut heuristics and language-surface bias, are controlled for by conducting evaluations across external datasets and non-English languages. Models trained with ESMA on English TriviaQA are evaluated without further training.

These rigorous controls consistently support that ESMA-driven improvements reflect a robust and generalized metacognitive capability.
A parameter patching analysis reveals that these behavioral gains are driven by a sparse subset of weight updates, suggesting the possibility of a specialized subnetwork linked to metacognitive functions.
The primary contributions of this paper are as follows:

\begin{itemize}
    \item Establishing a bias-controlled framework using the $d'_{\rm type2}$ metric to isolate metacognitive functions from response tendencies. This approach addresses multiple confounding factors, mitigating the illusion of knowing by checking consistency across dual-question and IDK prompts and overcoming the cue-familiarity heuristic via evaluation on a fictional dataset.
    \item Proposing ESMA to address limitations of gradient-based learning. This strategy evaluates a population of perturbed models using joint rewards to optimize metacognitive alignment by enforcing the consistent verification of knowledge across independent inference passes.
    \item Conducting a parameter patching analysis attributing metacognitive improvement to a sparse subset of high-magnitude weight updates. This raises the possibility of a specialized metacognitive subnetwork and motivates further exploration of targeted optimization.
\end{itemize}

For a detailed discussion of the psychological foundations of metacognitive measurement and their implications for the choice of tasks, metrics, and experimental controls, please refer to \cref{subsec:considerations}.

\section{Related Work}
\label{sec:related-works}

\begin{figure*}[t!]
    \centerline{\includegraphics[width=\textwidth]{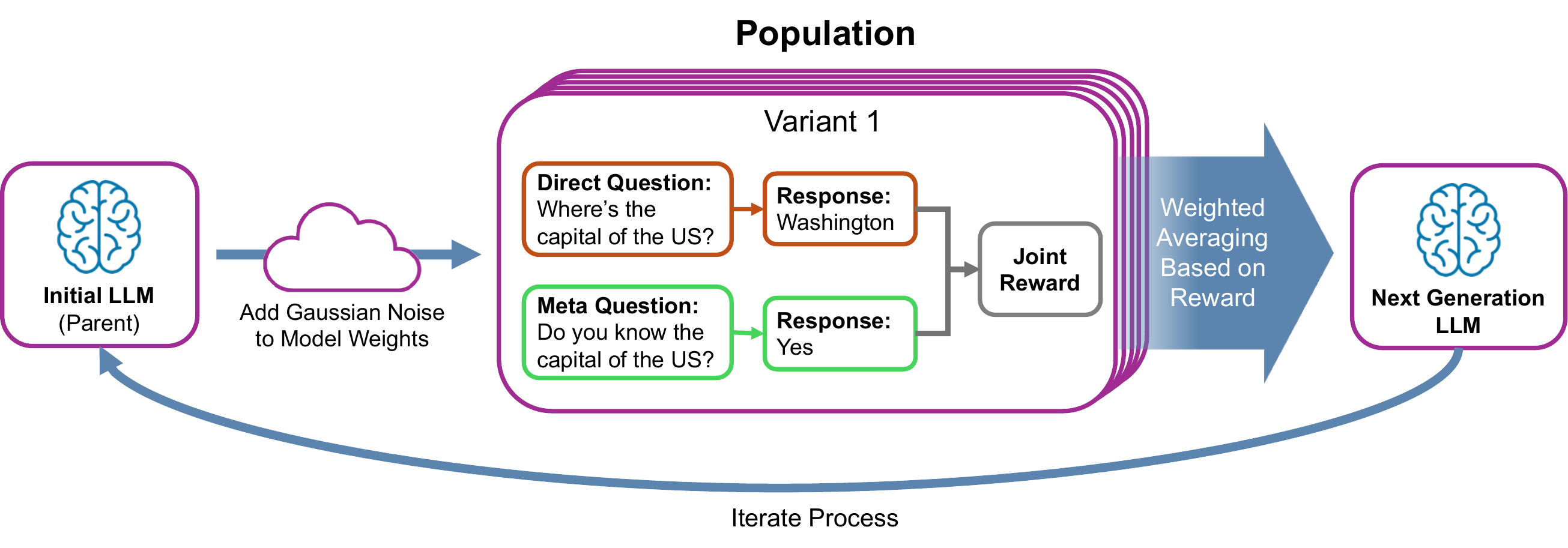}}
    \caption{Overview of Evolution Strategy for Metacognitive Alignment (ESMA). The process begins with an initial parent LLM, whose weights are perturbed with Gaussian noise to create a population of model variants. Each variant is evaluated on paired questions: a \textit{direct question} (to test factual knowledge) and a \textit{meta question} (to test self-knowledge). A joint reward is calculated based on the alignment between the correctness and meta answers, measuring whether the meta answer can reliably predict its correctness. Finally, the next-generation LLM is produced by a weighted average of the variants, prioritizing higher rewards. This process repeats for the specified number of generations.}
    \label{fig:training-process}
\end{figure*}

\paragraph{Metacognition in LLMs} 

The question of whether LLMs possess genuine metacognitive abilities remains a subject of ongoing debate \citep{ferrando2025do}.
Some scholars report a lack of metacognition in LLMs \citep{Yin2023-rb, Kale2025-ad, doi:10.1177/09637214251391158, singh2026can}, and others suggest models possess functional access to internal states \cite{comsa2025doesmakesensespeak, ji-an2026language, lindsey2025emergent, ackerman2026evidence}.
While studies on calibration \citep{Kadavath2022-ce}, refusal \citep{Prato2024-bf}, and behavioral prediction \citep{binder2025looking, Betley2025-cb} are often associated with metacognitive traits, they primarily address functional reliability and do not explicitly claim to investigate metacognition. Since research directly targeting metacognition remains limited \citep{decoupling-metacog-quantify}, this study focuses on the capacity to evaluate internal knowledge by controlling for confounding heuristics and proposes a method to advance this metacognitive alignment.

\paragraph{Evolution Strategies}
Evolution Strategies \citep[ES;][]{rechenberg1973evolutionsstrategie, schwefel1977} are gradient-free optimization techniques that iteratively refine model parameters by evaluating a population of perturbed candidates \citep{salimans2017evolution}.
Unlike backpropagation, which computes exact derivatives, ES requires only forward passes.
Recent advancements successfully extended this approach to the full-parameter fine-tuning of LLMs \cite{qiu2025, sarkar2025evolution}.

ES remains effective even for non-differentiable objectives where gradient-based approaches typically fail.
Since ES generates individuals with updated weights prior to reward calculation, each candidate can be evaluated through its entire behavioral patterns across diverse contexts rather than isolated actions.
The alignment between meta answers to self-knowledge and factual correctness is one such holistic objective, making ES particularly well suited to optimizing metacognitive alignment.

\section{Measurement Framework}
\label{sec:measurement}
Measuring human metacognition is operationalized via confidence-accuracy paradigms \cite{KUNIMOTO2001294}, where subjects solve a cognitive task and judge their accuracy. Drawing on these psychological paradigms, this work establishes a framework to measure $d'_{\rm type2}$ in LLMs.
For details regarding the measurement of human metacognition and the proposed extension, refer to \cref{subsec:measuring-human-metacognition}.

\paragraph{Task Setup}

Adapted from human metacognitive assessment frameworks, a dual-questioning protocol is implemented to evaluate LLMs. For each data point, models address two tasks: (1) \textit{direct questions} to elicit factual answers, and (2) \textit{meta questions} to evaluate self-knowledge. Isolating these inquiries within independent contexts ensures that meta answers preclude self-confirmation biases that could act as confounding variables \citep{Koriat2005-md}.

\paragraph{Metric}

$d'_{\rm type2}$ is a standardized metric that quantifies metacognitive ability. It measures the capacity to discriminate between accurate and inaccurate judgments by calculating the distance between the internal confidence distributions for each decision type.
\begin{equation}
d'_{\rm type2} = \Phi^{-1}(\text{Hit Rate}) - \Phi^{-1}(\text{False Alarm Rate}),
\end{equation}
This formula quantifies metacognitive performance as the statistical separation between distributions of correct and incorrect questions. A higher value indicates that the internal confidence reliably predicts accuracy.

In signal detection theory \cite{sdt-d-prime}, $d'_{\rm type2}$ values reflect discrimination sensitivity, ranging from chance-level ($d'_{\rm type2} \approx 0$) to an effective ceiling of near-perfect performance ($d'_{\rm type2} \ge 2.5$). Intermediate values around 0.5 and 1.0 indicate low (typical lower bound) and moderate (standard) sensitivity, respectively.

While the $d'_{\rm type2}$ metric is rigorous, it is helpful to also track other metrics that are more intuitive:
\begin{itemize}
    \item Raw Alignment: The averaged alignment between the meta answer and the correctness of the direct outcome. 
    While this measure is intuitive, it is susceptible to biases and statistical coincidences.
    For instance, a model that indiscriminately answers \texttt{``No''} would achieve a higher score in the case where questions are highly difficult overall.
    \item Accuracy: The ratio of correct answers provided for direct questions.
    \item Yes Ratio: The frequency with which the model provides a positive response to meta questions, indicating a belief that it knows the answer.
    \item Yes Failure Ratio (YFR): The proportion of incorrect direct answers among questions for which the model claims to know the answer.
    \item No Failure Ratio (NFR): The proportion of correct direct answers among questions for which the model claims not to know the answer.
\end{itemize}

\section{Evolution Strategy for Metacognitive Alignment}
\label{sec:evolutionary-training}

Optimizing for metacognitive alignment inherently requires consistent verification of knowledge across diverse situations, challenging standard gradient-based learning. Since backpropagation cannot compute gradients across independent inference passes, ES is used to design Evolution Strategy for Metacognitive Alignment (ESMA). Please refer to Appendices \ref{sec:evolution-strategies} and \ref{subsec:sft} for further details on ES and comparison with other baselines such as supervised fine-tuning and reinforcement learning (RL).

\begin{algorithm}
\caption{Evolution Strategy for Metacognitive Alignment}
\begin{algorithmic}[1]
\STATE \textbf{Initialize:} Parameters $\theta_0$, learning rate $\alpha$, mutation strength $\sigma$
\FOR{generation $t = 0, 1, 2, \dots, T-1$}
    \STATE Sample noise vectors $\epsilon_1, \dots, \epsilon_N \sim \mathcal{N}(0, I)$
    \FOR{each individual $i = 1, \dots, N$}
        \STATE Create perturbed model: $\theta_{i} = \theta_t + \sigma \epsilon_i$
        \STATE Evaluate fitness $F_i$ based on joint reward (Equation \ref{eq:joint_reward})
    \ENDFOR
    \STATE Obtain mean $\mu_F$ and standard deviation $\sigma_F$ of $\{F_1, \dots, F_N\}$
    \STATE Z-standardize fitness: $\hat{F}_i = \frac{F_i - \mu_F}{\sigma_F}$ for all $i$
    \STATE Update parameters: $\theta_{t+1} = \theta_t + \alpha \frac{1}{N} \sum_{i=1}^{N} \hat{F}_i \epsilon_i$
\ENDFOR
\end{algorithmic}
\label{alg:es-metacognition-ft}
\end{algorithm}

As illustrated in \cref{fig:training-process}, ESMA generates a candidate population by adding Gaussian noise to the model parameters. Each candidate is evaluated via a reward function, and the model is updated by calculating a weighted average of the perturbations, effectively moving the distribution toward regions that yielded the highest rewards.
Through iterative generations, this process progressively explores the parameter space for configurations that jointly optimize correctness and metacognitive alignment. The complete method is summarized in \cref{alg:es-metacognition-ft}.

\paragraph{Reward Design}

The reward function is designed along two primary axes: the direct correctness of the model's answer and its metacognitive alignment. Optimizing solely for direct correctness leaves metacognitive alignment untrained, whereas optimizing only for metacognitive alignment invites reward hacking. A model may maximize rewards by intentionally providing incorrect answers while consistently claiming ignorance. Please refer to \cref{subsec:univariate-reward-function} for experimental validation.

To prevent such degenerate behaviors, a differential reward system is applied based on the joint outcome of the direct and meta questions. Let $C=1$ if the direct answer is correct and $C=0$ otherwise. Let $A=1$ if the meta answer is aligned with the actual knowledge state (e.g., saying \texttt{Yes} when $C=1$, or \texttt{No} when $C=0$) and $A=0$ otherwise. The joint reward is used as follows:
\begin{equation}
    \label{eq:joint_reward}
    R(C, A) = 
\begin{cases} 
2 & \text{if } C=1, A=1 \text{ (Correct \& Yes)} \\
1 & \text{if } C=1, A=0 \text{ (Correct \& No)} \\
1 & \text{if } C=0, A=1 \text{ (Incorrect \& No)} \\
0 & \text{if } C=0, A=0 \text{ (Incorrect \& Yes)}
\end{cases}
\end{equation}
This formulation acts as a merged reward that unifies correctness and alignment into a single signal, preventing reward hacking by design. In particular, (Correct \& No) and (Incorrect \& No) both yield the same reward, removing any incentive to intentionally sacrifice correctness. Accordingly, the correctness term is included only to the extent necessary to preserve existing knowledge.

\section{Experiments}
\label{sec:experiments}
\begin{table*}[t!]
    \centering
    \small
    \setlength{\tabcolsep}{4pt}
    \caption{Bias-controlled evaluation of metacognition across proprietary and open-source LLMs.
Alongside raw alignment and behavioral metrics, $d'_{\rm type2}$ measures how well meta answers
discriminate correct from incorrect direct answers while controlling for global response tendencies.
The proprietary models illustrate the limitation of raw alignment: high task accuracy combined with
a strong positive response tendency can yield high apparent alignment without correspondingly high
metacognitive sensitivity. Across all open-source models, ESMA substantially increases
$d'_{\rm type2}$, showing that it improves metacognitive discrimination beyond a simple
shift in the overall response strategy.}
    \label{tab:overall-result}
    \begin{tabular}{lrrrrrr}
        \toprule
        Model & $d'_{\rm type2} \uparrow$ & Raw Alignment $\uparrow$ & Accuracy $\uparrow$ & Yes Ratio & YFR $\downarrow$ & NFR $\downarrow$ \\
        \midrule
        OpenAI GPT 5.2 & 0.94 & 84.66\% & 85.24\% & 92.59\% & 12.25\% & \textbf{53.95\%} \\
        Gemini 3 Flash & 0.68 & \textbf{90.71\%} & 92.68\% & 96.34\% & 6.72\% & 76.98\% \\
        Claude Sonnet 4.5 & \textbf{0.95} & 81.87\% & \textbf{93.19\%} & 82.20\% & \textbf{4.35\%} & 81.81\% \\
        \midrule
        Qwen2.5 1.5B       & 0.20 & 53.30\% & \textbf{42.86\%} & 53.81\% & 53.56\% & 38.69\% \\
        Qwen2.5 1.5B ESMA & \textbf{0.93} & \textbf{68.86\%} & 41.86\% & 37.89\% & \textbf{35.86\%} & \textbf{28.26\%} \\
        \midrule[0.05pt]
        Qwen2.5 3B         & 0.29 & 62.70\% & 35.67\% & 18.22\% & 54.46\% & 33.47\% \\
        Qwen2.5 3B ESMA   & \textbf{1.02} & \textbf{69.59\%} & \textbf{51.20\%} & 51.41\% & \textbf{29.78\%} & \textbf{31.07\%} \\
        \midrule[0.05pt]
        Qwen2.5 7B         & 0.64 & 61.65\% & 50.43\% & 36.20\% & 33.31\% & 41.21\% \\
        Qwen2.5 7B ESMA   & \textbf{0.94} & \textbf{69.86\%} & \textbf{60.71\%} & 67.64\% & \textbf{27.40\%} & \textbf{35.87\%} \\
        \midrule[0.05pt]
        Gemma3 4B         & 0.04 & 52.75\% & 46.53\% & 19.12\% & 51.88\% & 46.15\% \\
        Gemma3 4B ESMA   & \textbf{0.92} & \textbf{67.94\%} & \textbf{55.58\%} & 53.29\% & \textbf{27.92\%} & \textbf{36.77\%} \\
        \midrule[0.05pt]
        Llama3.2 3B         & 0.20 & 49.57\% & 53.74\% & 14.60\% & 38.66\% & 52.44\% \\
        Llama3.2 3B ESMA   & \textbf{0.89} & \textbf{67.73\%} & \textbf{55.64\%} & 58.10\% & \textbf{29.89\%} & \textbf{35.58\%} \\
        \bottomrule
    \end{tabular}
\end{table*}

The experiments are designed to evaluate metacognitive ability under bias-controlled conditions and to test whether ESMA produces improvements that cannot be explained by alternative heuristic-driven strategies.
The study utilized open-source instruct-tuned models including Qwen2.5, Llama3.2, and Gemma3 in various sizes \cite{qwen2025qwen25technicalreport, llama3, gemma3}.
For comparison, proprietary models such as OpenAI GPT 5.2 \cite{gpt5}, Claude Sonnet 4.5 \cite{claude45}, and Gemini 3 Flash \cite{gemini3flash2025} were also evaluated.
Refer to \cref{sec:experimental-details} for specific hyperparameters and prompts.

\paragraph{Dataset}
To isolate metacognition from confounding variables like linguistic fluency or complex reasoning, TriviaQA \cite{joshi-etal-2017-triviaqa} was selected as the training source and primary benchmark.
TriviaQA is particularly suitable as it comprises factual questions with short keyword answers (e.g., ``Which city does David Soul come from?'').
By limiting the task demands to simple knowledge retrieval, the model's ability to recognize the presence of specific information is more effectively isolated.

\subsection{Metacognitive Ability}

\cref{tab:overall-result} presents the comprehensive results of various models, revealing critical trends in LLM metacognition and the efficacy of the ESMA method.
First, the proprietary results demonstrate why $d'_{\rm type2}$ is essential for a bias-controlled metric of metacognition. These models achieve very high direct-question accuracy, ranging from 85.24\% to 93.19\%, and consequently obtain high raw alignment scores. However, raw alignment can be inflated when high task accuracy coincides with an overconfident meta response tendency. For example, if a model answers 90\% of direct questions correctly, it can achieve 90\% raw alignment simply by answering \texttt{Yes} to every meta question without any item-level information.

Gemini 3 Flash achieves the highest raw alignment among the proprietary models at 90.71\%, but its $d'_{\rm type2}$ is the lowest among them at 0.68. This discrepancy suggests that Gemini's high raw alignment is driven primarily by the match between its high direct accuracy and strong positive response tendency, as reflected in its yes ratio of 96.34\%. In contrast, $d'_{\rm type2}$ penalizes such false alarms and therefore better isolates the statistical predictive power of the meta answer. The closed-model comparison thus validates the role of $d'_{\rm type2}$ as a bias-controlled metric of metacognitive sensitivity.

In contrast, open-source LLMs exhibited relatively weak baseline metacognitive performance. Their low $d'_{\rm type2}$ scores indicate that their meta answers were only weakly informative about the correctness of their direct answers before fine-tuning with ESMA.

The application of ESMA produced significant improvements across all open-source models. For every model, ESMA caused a sharp increase in $d'_{\rm type2}$, elevating scores toward the moderate level of 1.0, accompanied by a consistent rise in raw alignment. Notably, Qwen2.5 3B ESMA achieved the highest $d'_{\rm type2}$ score of 1.02 across all evaluated models, surpassing even the proprietary counterparts despite having substantially lower direct-question accuracy. This result suggests that ESMA improves metacognitive discrimination, rather than merely inducing a global response-strategy shift.

Within a similar parameter class, Qwen2.5 3B outperformed both Gemma3 4B and Llama3.2 3B.
Also, the Qwen series shows a positive correlation between model size and metacognitive ability.
This implies that while metacognitive behavior naturally emerges alongside the generalization of models, ESMA consistently propels performance beyond the natural baselines across all model sizes.

Appendices \ref{subsec:external-datasets-evaluation} and \ref{subsec:cross-lingual-evaluation} provide external and cross-lingual results confirming ESMA's efficacy.
\cref{sec:qualitative-examples} shows how ESMA qualitatively changes the behavior of the model.

\begin{figure*}[t!]
    \centerline{\includegraphics[width=0.95\textwidth]{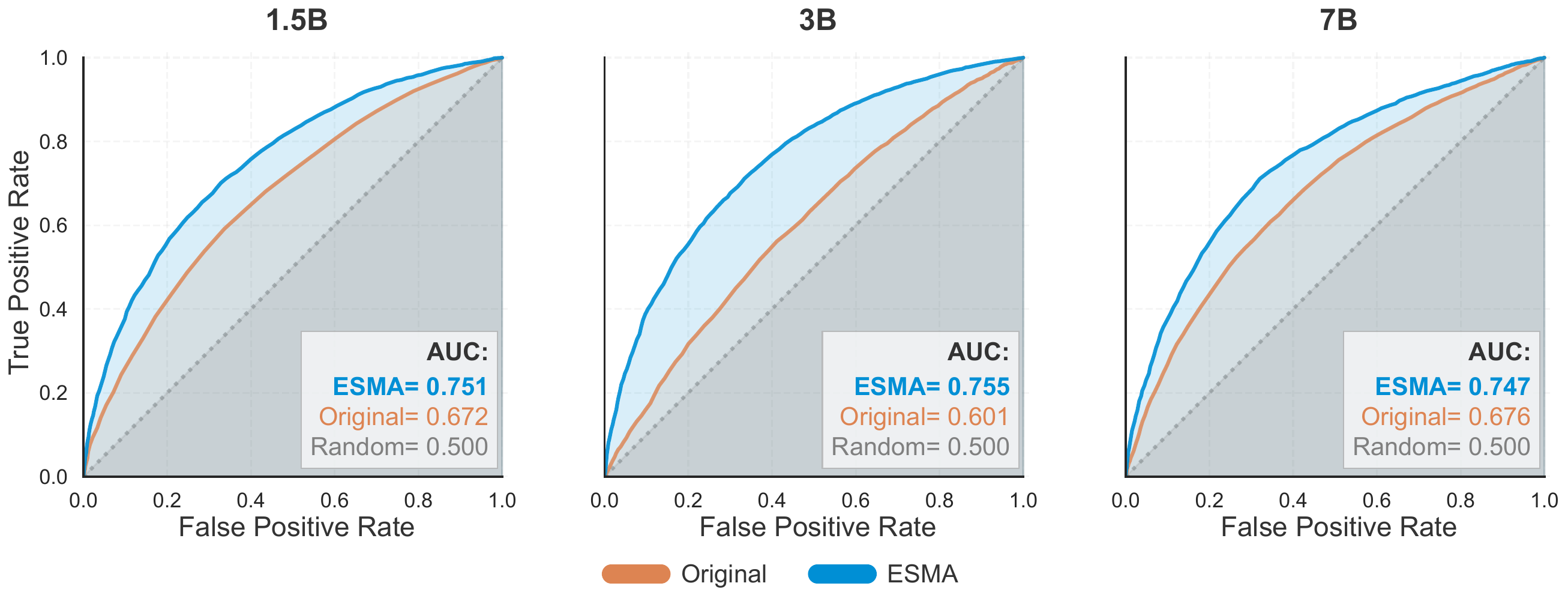}}
    \caption{Type 2 ROC curves using continuous confidence scores. As binary responses can be sensitive to local fluctuations near the \texttt{Yes/No} decision boundary, Type 2 AUROC provides a complementary continuous check by measuring how well confidence scores rank correct direct answers above incorrect direct answers across thresholds. ESMA consistently increases AUROC across all scales, showing that its improvement is also reflected in the continuous confidence.
    }
    \label{fig:type2-auc-roc}
\end{figure*}

\subsection{Confidence-based Metacognition}
\label{subsec:confidence-based-metacognition}

The main evaluation uses discrete meta answers and $d'_{\rm type2}$ as the measure of metacognitive sensitivity. This setup is useful because the binary response reduces ambiguity in self-reports and avoids scale-use artifacts often observed in graded confidence ratings, such as central-tendency bias \citep{Xiang2021-wj, Benjamin2013-qu}.

Nevertheless, discrete answers provide a coarse observation of an underlying confidence signal. Small sampling or logit-level fluctuations near the \texttt{Yes/No} decision boundary may affect the recorded binary response. Therefore, continuous Type 2 AUROC provides a complementary metric by measuring how well the model ranks correct direct answers above incorrect direct answers by confidence, across all possible thresholds.

For this purpose, confidence $D$ was calculated as the normalized probability of a positive self-assessment, derived from the logits of the \texttt{Yes} and \texttt{No} tokens.
\begin{equation}
D = \frac{P(\texttt{Yes})}{P(\texttt{Yes}) + P(\texttt{No})} .
\end{equation}
Psychology research frequently utilizes Type 2 AUROC for granular metacognitive ratings \citep{Fleming2010-ex}. In alignment with this standard, the analysis applied a Receiver Operating Characteristic (ROC) analysis.
The ROC curve traces the True Positive Rate (TPR) against the False Positive Rate (FPR) as the confidence threshold is swept from 0 to 1.
The resulting AUROC summarizes the model's ranking ability over the full continuous confidence scale.
It can be interpreted as the probability that a randomly selected correct direct answer receives a higher confidence score than a randomly selected incorrect direct answer.

Thus, AUROC values are interpreted as follows: 0.5 represents a random baseline, 0.5–0.7 indicates low to marginal metacognitive ability, and values exceeding 0.7 reflect moderate to high ability to reliably distinguish between known and unknown information.

Sweeping the confidence threshold from 0 to 1 produced the ROC trajectories for each model. As illustrated in \cref{fig:type2-auc-roc}, the original models already exhibited AUROC values above the 0.5 random baseline. This result indicates that the continuous confidence scores of the original models contain weak but nonzero information about answer correctness, even before explicit metacognitive tuning.

The ROC trajectories also reveal differences across the original models. The original 3B model showed a particularly notable plateau in the low-FPR region, remaining relatively low in TPR until the false positive rate approached 0.2. This pattern suggests that, compared with the 1.5B and 7B models, the original 3B model more often assigned high confidence to incorrect answers, making its continuous confidence signal less reliable at the strictest thresholds. This observation is consistent with its lower original AUROC of 0.601.

After ESMA, the ROC curves shifted consistently toward the upper-left corner across all model scales, significantly increasing the AUROC for all models. Thus, all ESMA models reached an AUROC of approximately 0.75, indicating substantially stronger ranking of correct direct answers above incorrect direct answers by confidence.

These results show that the gains observed with $d'_{\rm type2}$ are not limited to the discrete decisions used in the main evaluation. Although ESMA was trained only through binary meta answers, its effect also appears in the continuous confidence signal. This provides additional evidence that ESMA improves metacognitive discrimination rather than producing a fragile artifact of binary response measurement.

\begin{figure*}[t!]
    \centerline{\includegraphics[width=0.95\textwidth]{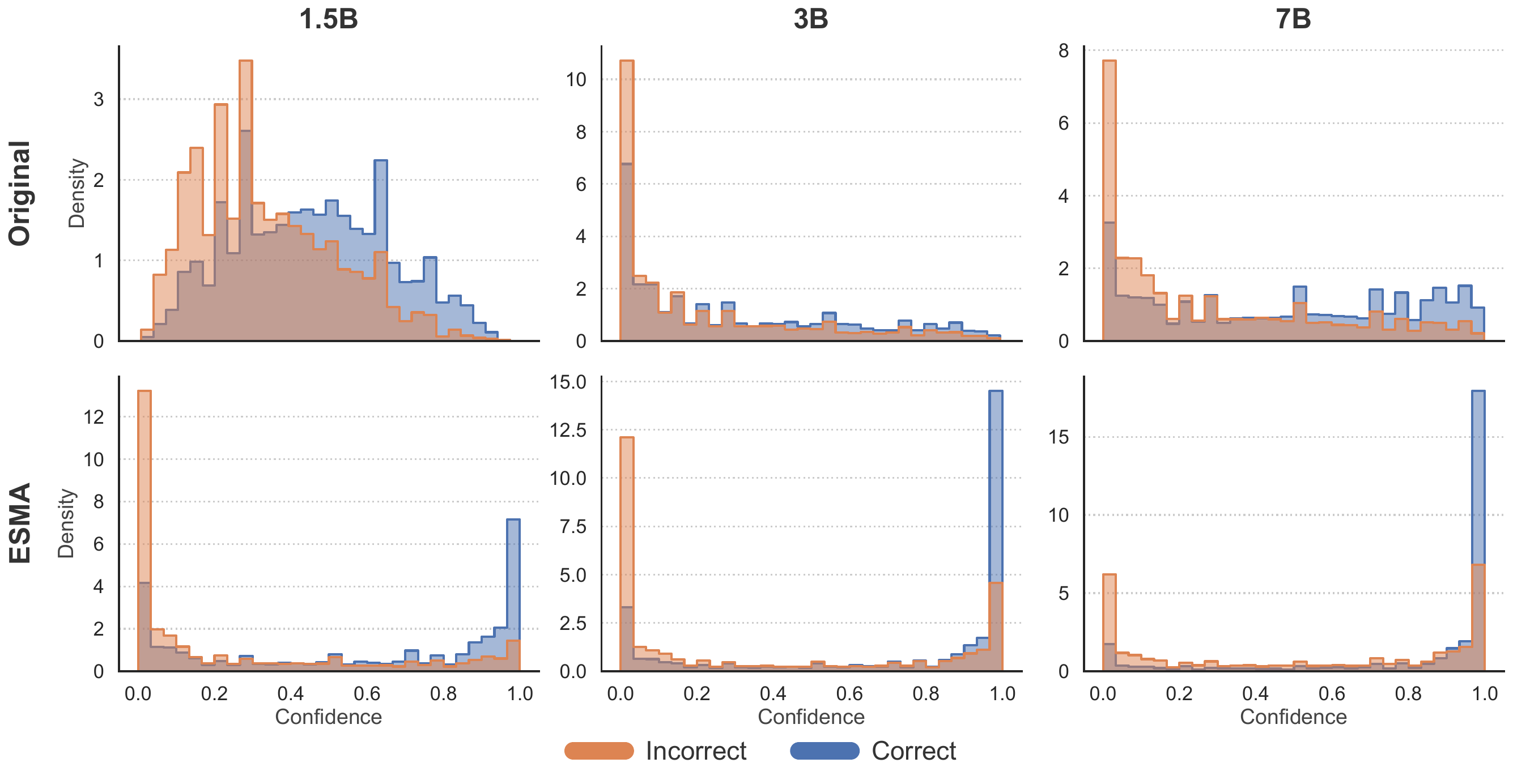}}
    \caption{
    Distribution of continuous metacognitive confidence by correctness. Density plots show the confidence score $D$ for correct and incorrect direct answers in the original and ESMA versions of the Qwen models. The original models have substantially overlapping distributions, indicating that correct and incorrect answers often receive similar confidence. After applying ESMA, confidence shifts toward 1.0 for correct answers and 0.0 for incorrect ones, reducing the overlap between the two distributions. This distribution-level separation supports the $d'_{\rm type2}$ and AUROC results.
    }
    \label{fig:confidence-distribution}
\end{figure*}

\subsection{Visualization of Confidence Distribution}

The Type 2 AUROC results show that ESMA improves the continuous ranking of correct and incorrect answers by confidence. To better understand this improvement, the confidence distributions for correct and incorrect instances were visualized directly. If the AUROC gain were caused only by local fluctuations around a binary decision boundary, the overall shapes of the two distributions would remain largely similar. In contrast, an improvement in metacognitive discrimination should appear as a broader separation between the confidence assigned to correct and incorrect answers.

As shown in \cref{fig:confidence-distribution}, the original models exhibited substantial overlap between the two distributions. In the 1.5B and 3B models, confidence scores were concentrated in a relatively ambiguous middle region, indicating that the models often assigned similar confidence to correct and incorrect answers. The 7B model showed a clearer baseline separation, consistent with the stronger metacognition observed in the previous analyses, but the two distributions still remained visibly overlapping.

ESMA changed the shape of these distributions across all model scales. Confidence for correct answers shifted toward 1.0, while confidence for incorrect answers shifted toward 0.0, producing a more polarized and better separated distributional structure. As a result, the overlapping region between correct and incorrect instances was substantially reduced. This pattern supports the interpretation of the AUROC results: ESMA does not merely alter a small set of binary decisions near the response boundary, but reshapes the continuous confidence signal so that correct and incorrect answers become more separable across the full confidence range.

This distribution-level change provides additional evidence that the improvements in $d'_{\rm type2}$ and Type 2 AUROC reflect stronger metacognitive discrimination.

\begin{table*}[t!]
    \centering
    \small
    \setlength{\tabcolsep}{4pt}
    \caption{Metacognitive ability under a unified prompt format where the model is permitted to respond \textit{I don't know} (IDK) instead of providing an answer. IDK accuracy reports task performance under this setting, while IDK alignment treats an abstention as a \texttt{No} meta answer and any other output as a \texttt{Yes} meta answer. All alignment is a cross-format consistency metric requiring that the dual-question correctness, dual-question meta answer, and IDK response all agree. Dual-question accuracy reproduces the accuracy values from \cref{tab:overall-result} for comparison. The bolded results address both bias concerns: ESMA's improvement in IDK alignment shows that its gains are not tied to the training-time prompt template, and its improvement in all alignment indicates that the enhanced meta answers remain consistent with actual correctness, mitigating the concern of an illusion of knowing.
    }
    \label{table:idk-alignment}
    \begin{tabular}{lrrrr}
        \toprule
        Model & Dual-question Accuracy $\uparrow$ & IDK Accuracy $\uparrow$ & IDK Alignment $\uparrow$ & All Alignment $\uparrow$ \\
        \midrule
        Qwen2.5 1.5B       & \textbf{42.86\%} & \textbf{38.21\%} & 52.34\% & 31.45\% \\ 
        Qwen2.5 1.5B ESMA & 41.86\% & 35.54\% & \textbf{64.63\%} & \textbf{45.94\%} 
        \\ 
        \midrule[0.1pt]
        Qwen2.5 3B         & 35.67\% & 35.38\% & 59.88\% & 28.87\% \\ 
        Qwen2.5 3B ESMA   & \textbf{51.20\%} & \textbf{40.88\%} & \textbf{78.07\%} & \textbf{59.71\%} \\ 
        \midrule[0.1pt]
        Qwen2.5 7B         & 50.43\% & 49.53\% & 60.05\% & 34.93\% \\ 
        Qwen2.5 7B ESMA   & \textbf{60.71\%} & \textbf{56.89\%} & \textbf{66.77\%} & \textbf{55.49\%} \\ 
        \bottomrule
    \end{tabular}
\end{table*}

\subsection{Metacognitive Alignment with a Unified Prompt}
\label{subsec:alignment-with-unified-prompt}

The \textit{I don't know} (IDK) experiment targets two remaining bias concerns in the dual-question evaluation: prompt-template bias and the illusion of knowing (IoK). First, because ESMA is trained with dual prompts, improved performance could reflect adaptation to that specific prompt template and the setup.
Second, as discussed in \cref{subsec:considerations}, IoK can make subjective knowledge reports appear valid even when they do not reliably track actual knowledge. Psychological studies therefore often examine the same knowledge state through multiple elicitation formats and test whether metacognitive judgments remain consistent across them.

\paragraph{Experimental Design}
The dual-question format is replaced with a single prompt by adding the instruction \texttt{``If you don't know the answer, just return `I don't know'{''}} to the direct question prompt.

In this IDK setup, the separate meta question is omitted. Instead, the model's output is considered a merged response: an answer of \texttt{``I don't know''} is categorized as a meta \texttt{``No''}, while any other response is categorized as a meta \texttt{``Yes''}. 
If this ability stems from accessing and referencing its internal knowledge states, rather than a simple pattern-matching response to specific meta questions, the model should retain this ability when queried using this setup.

Critically, this prompt was never used during the fine-tuning phase, and the models were never explicitly optimized to respond \texttt{``I don't know''}. Consequently, the performance in this setting acts as a zero-shot probe into whether ESMA established a robust, format-independent association between internal knowledge and metacognitive response, rather than relying on format-specific heuristics or prompt-dependent biases.

\paragraph{Metrics}
Three primary metrics were defined to assess performance in this setting:
\begin{itemize}
    \item \textbf{IDK Accuracy}: The accuracy of the model on the direct question within this integrated prompt setting.
    \item \textbf{IDK Alignment}: The metacognitive alignment in this setting, calculated by comparing correctness with the occurrence of an \texttt{``I don't know''} response.
    \item \textbf{All Alignment}: A stringent consistency metric. An instance is considered fully aligned only if the correctness and the meta answer in a dual-question setting, and the current setting's meta answer are all mutually consistent. This measures the stability of the model's internal epistemic signal across prompt formats.
\end{itemize}

\paragraph{Results and Discussion}
The results of this experiment are summarized in \cref{table:idk-alignment}. As anticipated, IDK accuracy was consistently lower than dual-question accuracy across all models, because allowing the model to refuse to answer uncertain questions inherently counts against its direct accuracy score. The findings revealed a significant improvement in the fine-tuned models. Crucially, the IDK alignment improved significantly across all scales (e.g., increasing from 59.88\% to 78.07\% on the 3B model) despite the models never being fine-tuned on this specific IDK prompt. This zero-shot improvement demonstrates that ESMA's gains are not merely the result of prompt-template bias or overfitting to the dual-question training format.  

Furthermore, the all alignment metric improved substantially, particularly in the 3B and 7B variants. By maintaining alignment across entirely different elicitation formats, this cross-format consistency directly mitigates the concern that the model is merely experiencing an illusion of knowing. Together, these results suggest that ESMA successfully develops a more robust metacognitive alignment, rather than relying on prompt-template bias or IoK.  

\subsection{Metacognition on New Information}

\begin{wraptable}{r}{0.5\textwidth}
    \vspace{-1em}
    \centering
    \small
    \setlength{\tabcolsep}{5pt}
    \caption{Performance on FictionalQA. ESMA significantly increased $d'_{\rm type2}$ to 0.65 despite no prior exposure to meta questions about the fictions, demonstrating that the model references its internal states for metacognitive judgments.}
    \label{table:fictional-qa-performance}
    \begin{tabular}{lrrr}
        \toprule
        Model & $d'_{\rm type2} \uparrow$ & Raw Al. $\uparrow$ & Acc. $\uparrow$\\
        \midrule
        Original        & 0.23 & \textbf{76.18\%} & 1.45\% \\
        + FictionalQA  & 0.20 & 53.92\% & \textbf{51.64\%}\\
        + ESMA  & \textbf{0.65} & {48.97\%} & 51.09\% \\
        \bottomrule
    \end{tabular}
    \vspace{1em}
\end{wraptable}

While previous experiments demonstrate ESMA's effectiveness in improving metacognition, a potential concern is that the model may rely on the cue-familiarity heuristic. 
If a model was exposed to similar meta question templates during the pretraining phase (e.g., ``Q: Do you know the capital of the US? A: Yes.''), its apparent metacognitive alignment might stem from recognizing familiar terms in the prompt rather than evaluating whether it can retrieve the correct answer.
To control for this bias, FictionalQA \cite{fictionalqa} was utilized. Released in June 2025, this dataset provides QA pairs about entirely fictional events that do not exist in real-world knowledge, ensuring that any pre-existing semantic familiarity cannot serve as a reliable proxy for answer correctness.

The Qwen2.5 1.5B model was fine-tuned on half of the direct QA pairs from FictionalQA using supervised fine-tuning (SFT), followed by the application of ESMA using the TriviaQA training set. \cref{table:fictional-qa-performance} presents the results. The nonzero accuracy of the original model was attributed to accidental matches with gold answers containing generic keywords. Accuracy increased to approximately 50\% after SFT. Remarkably, ESMA resulted in a significant increase in $d'_{\rm type2}$ to 0.65. Given that the model had never been exposed to the meta question prompts in FictionalQA, this result suggests that ESMA establishes a functional link between the model's self-reports and its newly acquired knowledge, independent of the cue-familiarity heuristic.

\subsection{Sparse Contribution of Evolutionary Weight Updates}

ES modifies all model parameters simultaneously at each iteration. However, even when a specific modification improves fitness, it does not necessarily follow that every individual parameter change contributes positively to the fitness. To investigate the efficiency and effectiveness of these learned updates, this study examined whether the improvements were driven by the collective shift of all parameters or by a subset of highly influential weight changes.

\paragraph{Experimental Procedure}
The total parameter change, $\Delta \mathbf{W}$, is defined as the difference between the weights of the final fine-tuned model, $\mathbf{W}_{\rm tuned}$, and the original model, $\mathbf{W}_{\rm original}$, as
\begin{equation}
    \Delta \mathbf{W} = \mathbf{W}_{\rm tuned} - \mathbf{W}_{\rm original} .
\end{equation}
To evaluate the impact of these changes, every individual scalar value within $\Delta \mathbf{W}$ was ranked by its magnitude (L1 norm). A patching ratio $p \in [0, 100]\%$ was then defined, representing the top $p\%$ of weight changes with the highest absolute values. Finally, a partially patched model, $\mathbf{W}_{p}$, was constructed by adding only these selected updates to the original model:
\begin{equation}    
    \mathbf{W}_{p} = \mathbf{W}_{\rm original} + \text{Top}_p(\Delta \mathbf{W}) ,
\end{equation}
where $\text{Top}_p(\Delta \mathbf{W})$ is a sparse tensor containing only the $p\%$ largest magnitude updates from $\Delta \mathbf{W}$ and zeros elsewhere. The parameter $p$ was increased from 0\% (the original model) to 100\% (the fully fine-tuned model) every 10\%, and the resulting $d'_{\rm type2}$ and raw alignment were measured.

\begin{wrapfigure}{r}{0.5\textwidth}
    \centering
    \vspace{-1em}
    \includegraphics[width=0.45\textwidth]{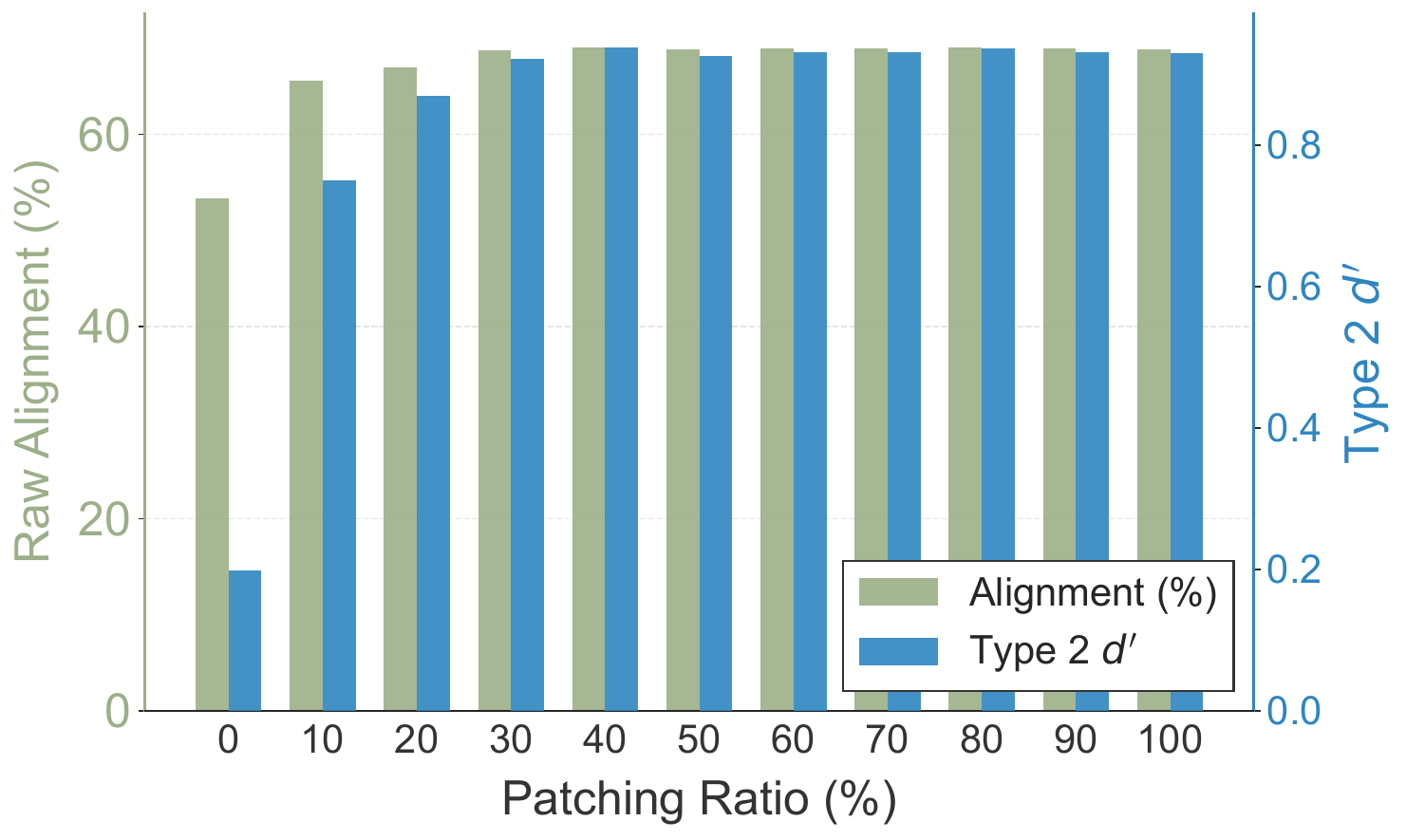}
    \caption{Effect of weight patching ratio on metacognitive abilities. The plot illustrates how $d'_{\rm type2}$ (blue bars, right axis) and raw alignment (\%, green bars, left axis) change as the top $p\%$ of parameter updates (by L1 magnitude) are applied to Qwen2.5 1.5B. Substantial gains were observed with patching only the top 10\% of parameters, after which performance plateaued. This observation suggests that metacognitive improvements can be achieved by training a sparse subset of parameters.}
    \label{fig:weight-change-effect}
    \vspace{-1em}
\end{wrapfigure}

\paragraph{Results}
As illustrated in \cref{fig:weight-change-effect}, the experiment revealed a highly non-linear relationship between the volume of weight patching and the improvement in metacognitive performance. The most striking observation is the efficiency of the highest-magnitude updates. Applying only the top 10\% of weight changes yielded a dramatic increase in performance: $d'_{\rm type2}$ rose sharply from 0.20 to 0.75, capturing roughly 75\% of the total improvement of the entire training.
Similarly, raw alignment increased from 53.30\% to 65.60\% in this initial 10\% window.

Beyond this initial surge, the rate of improvement slowed considerably. Increasing the patching ratio from 10\% to 40\% led to a more gradual improvement, with $d'_{\rm type2}$ peaking at 0.94 and raw alignment stabilizing around 69\%. Notably, the remaining 50\% of weight updates, consisting of the smallest scalar changes, contributed almost no additional metacognitive improvement.
The fact that the vast majority of improvement was concentrated in the top 10\% of parameter updates indicates that metacognitive alignment is driven by specific, high-impact weight adjustments rather than a uniform shift across the entire parameter space.
If a more essential parameter set that substantially drives these improvements can be isolated, it would present a promising opportunity to identify a specialized subnetwork linked to metacognitive functions in future research.
Please refer to \cref{subsec:detailed-analysis-weight-update} for additional analyses of the top weight change and a bottom-up patching experiment.

\section{Discussion and Future Work}
\label{sec:discussion}
This study demonstrates that experimental setups for measuring human metacognition can be applicable to LLMs, where statistical tools like $d'_{\rm type2}$ and Type 2 AUROC are crucial for separating internal knowledge access from response strategies. Still, measuring metacognition in LLMs is in its early stages. Metacognition is not limited to the simple fact retrieval task, and the self-monitoring skill required for advanced reasoning will differ from metacognition in the factual retrieval task.
Another important direction is to evaluate whether LLMs can accurately assess their ability to do something. For example, success in reasoning depends on a sequence of steps whose outcomes may be difficult to anticipate. This distinction becomes increasingly relevant for agentic LLMs that use external tools, where metacognition requires assessing not only what a model knows but also what it can accomplish.

Although ESMA improves metacognitive alignment of LLMs, their performance remains below human standards even in simple factual retrieval. A central contribution of this work is to operationalize metacognitive ability through a bias-controlled measure that can serve as an explicit optimization target. This operational definition provides a basis for systematically studying metacognitive capabilities in LLMs. Ultimately, the remaining gap calls for further advances to improve metacognition as an independent cognitive capability rather than merely a tool for correcting problematic behaviors.

An important foundation of human metacognition is the explicit memory system, which supports the recollection of semantic knowledge and episodic events. Semantic memory stores accessible general knowledge, while episodic memory links this knowledge to the contexts where it was acquired or utilized. Recalling these contexts can help humans assess the reliability of information based on its source or draw on past experiences of utilizing that knowledge. This suggests that achieving human-level metacognition in LLMs may require developing more sophisticated memory systems.

Finally, this work illustrates the distinctive potential of Evolution Strategies (ES) in modeling complex cognitive abilities. Unlike traditional methods that focus on independent input-output pairs, ES achieves metacognitive success through the direct optimization of comprehensive behavioral patterns beyond a single context. This finding highlights the potential of ES for optimizing higher functions that require holistic interactions. Extending ES to domains such as strategic reasoning will be advantageous, where effective behavior must be jointly evaluated across interdependent decisions.

\bibliography{main}
\bibliographystyle{plainnat}
\newpage


\appendix

\section{Measuring Metacognition}
\label{sec:measuring-metacognition}
\subsection{Motivation for the Study Design}
\label{subsec:considerations}

\paragraph{Caveats in Metacognitive Measurement}

The primary subject of exploration in this study, metacognitive knowledge \citep{1980-09388-001, NELSON1990125}, refers to awareness of one's own knowledge. While metacognition has long been an important subject in psychology, it remains a difficult topic because awareness is fundamentally unobservable in a direct manner. Nevertheless, an operational definition is necessary to conduct practical research. Therefore, the field of psychology uses the statistical correspondence between objective performance and subjective reports as the operational definition of metacognitive monitoring \citep{schraw2009, fleming2014}.

Although this operational definition has enabled the quantification of metacognition, challenges remain \citep{KATYAL2024223, Rahnev2025-vw}. The problem is that the statistical correspondence between subjective reports and objective performance can be influenced by various heuristics, independently of the actual awareness of internal states \citep{evans2007evaluation}. For example, since subjects possess individual response biases \citep{Masson2009-pi}, certain individuals can generally exhibit overconfidence in their knowledge. If the task happens to be easy, metacognition might be accidentally measured as high. Another example is the processing fluency heuristic \cite{Benjamin1998-bk, Alter2009-sr}. When the brain processes specific information easily and smoothly, it can mistakenly interpret this as a signal that the knowledge is well understood. \citet{Rhodes2008} also demonstrated in a word memorization task that when some words were presented in a large font and others in a small font, subjects reported high subjective confidence in remembering the larger words the next day. However, this confidence did not show a meaningful correlation with their actual performance.

The operation of such heuristics functions as a powerful confounding variable that influences measured values regardless of the true awareness of the knowledge itself, thereby hindering the objective of accurately measuring metacognitive ability. Consequently, controlling for heuristics and biases is an essential procedure in studies aiming to measure and compare metacognition \citep{MANISCALCO2012422, Bjork2012-jt}.

\paragraph{Metacognition in LLM Research}

There is a parallel challenge in LLM research. Research on metacognition in LLMs remains in its early stages but has attracted increasing attention \citep{NEURIPS2024_2318d75a, griot2025, ma-etal-2025-large-language-models}. Some researchers associate metacognition with output reliability evaluation or self-verification abilities \citep{wang-zhao-2024-metacognitive, liu-etal-2025-metafaith, li-etal-2025-adaptive, zhou2024metacognitive}. This association likely arises because issues such as hallucinations \citep{Ji_2023} or inappropriate refusals are often framed as metacognitive deficits \citep{lu2026auditing}. Consequently, the mitigation of hallucination errors is interpreted as an enhancement in metacognitive ability. However, even these studies typically do not claim that they improved metacognitive capacity, instead focusing on the functional reduction of errors as their primary achievement.

Interpreting performance improvements on these tasks as evidence of metacognition is problematic because, as in human studies, such outcomes can be driven by various heuristics rather than access to internal states \citep{xiong2024can, 10.5555/3666122.3669397}. These heuristics can even be beneficial in some cases. For example, assigning lower confidence to questions perceived as difficult may improve practical reliability without reflecting any self-knowledge. It means that the resulting metrics do not represent metacognitive ability in a strict sense.

This limitation is particularly evident in research areas such as hallucination mitigation, refusal behavior \citep{Zhang2024-es, wen-etal-2025-know}, and confidence calibration \citep{manakul-etal-2023-selfcheckgpt, shinn2023reflexion, lin2022teachingmodelsexpressuncertainty}. For example, Retrieval-Augmented Generation (RAG) can make answers more reliable \citep{10.5555/3495724.3496517}, reduce hallucinations \citep{ayala-bechard-2024-reducing}, and increase appropriate refusal for uncertain queries \citep{asai2023selfraglearningretrievegenerate}; however, these gains arise from augmenting the model with external information rather than altering its internal representations or its capacity to assess its own knowledge. Similarly, post hoc methods such as temperature scaling can reduce Expected Calibration Error (ECE) and improve confidence calibration \citep{pmlr-v70-guo17a}, but they do so without modifying the model's underlying parameters or inference mechanisms, and thus cannot constitute metacognitive change.

Consequently, results observed in these tasks are difficult to interpret as being directly related to metacognitive ability or epistemic access. Rather, they are more appropriately understood in terms of the specific problems they are designed to address, as framed in the original studies themselves. To rigorously investigate the measurement and improvement of metacognitive capacity, it is necessary to develop approaches that explicitly mitigate such confounding factors.

In addition, LLM research introduces model-specific evaluation confounds \citep{shortcut-learning} that are not typically central in human metacognition studies. Because LLM outputs are mostly prompt-conditioned and benchmark-dependent, apparent metacognitive improvements may arise from prompt-template bias \citep{zhuo-etal-2024-prosa}, benchmark-specific heuristics \citep{mccoy-etal-2019-right}, or language-surface bias \citep{qi-etal-2023-cross}. Thus, alongside traditional psychological biases and heuristics, these model-specific factors require additional consideration when studying metacognition in LLMs.

\paragraph{Rationale for Experimental Design}

The experimental design of this study was developed to measure metacognition rigorously in LLMs by directly mapping and mitigating the confounding factors identified in human psychology. 

First, to establish a foundation, the standard operational definition of metacognition used in psychological research was adopted: the statistical correspondence between objective performance and subjective reports. While LLM studies often focus on functional task performance as a proxy, the model is directly queried about its awareness to capture this subjective report in the dual-prompt setup. Simple question-answering tasks were specifically utilized to explore the model's internal accessibility to its self-knowledge. This setup isolates the knowledge retrieval phase, thereby precluding the influence of complex reasoning.

Second, simply asking a model about its awareness does not automatically guarantee genuine introspection. Querying an LLM makes it susceptible to response biases, such as overconfidence or random guessing, along with difficulty-based heuristics. In psychology, researchers recognize that relying on naive measures of agreement between accuracy and self-report is highly vulnerable to these biases. Instead, they employ statistical metrics like $d'_{\rm type2}$ or meta-$d'$ to quantify how well subjective reports discriminate between correct and incorrect responses, independent of a subject's response bias. In direct alignment with this necessity, this study adopts the $d'_{\rm type2}$ metric. This approach allows for a more robust evaluation of the model's metacognitive capacity by mitigating the influence of prompt-template priors or innate response tendencies.

Third, psychological research identifies phenomena like the processing fluency heuristic and the illusion of knowing (IoK) as major hurdles. To circumvent these, psychologists employ methodological strategies, such as multi-stage questioning to verify a single piece of knowledge from different angles \citep{Rozenblit2002-bd}, or delayed judgments of learning \citep{Avhustiuk2018-ks}. The current experimental setup purposefully mirrors these efforts. To avoid the illusion of knowing, independent meta questions were introduced to verify consistency against direct questions, and this alignment was further validated using an alternative IDK-based evaluation setting where uncertainty is expressed in a different format. Additionally, human subjects often rely on cue-familiarity heuristics \citep{Metcalfe1993-oq, Koriat2001}. To bypass this heuristic in LLMs, the FictionalQA experiment was designed to measure metacognitive monitoring on newly acquired, entirely fictional information.

Finally, the design additionally considers model-specific confounds in LLM evaluation. Beyond traditional psychological biases and heuristics, LLM self-reports may be affected by benchmark-specific shortcut heuristics or language-surface bias. The evaluation on external datasets and cross-lingual experiments therefore serve as complementary checks that the observed alignment is not merely tied to benchmark-specific heuristics or language-surface bias.


To the best of our knowledge, this study is the first to explore LLM metacognitive abilities by systematically controlling for biases and alternative hypotheses using methodologies directly modeled after psychological research, representing one of the core contributions of this paper.

\subsection{Metrics}
\label{subsec:measuring-human-metacognition}

In psychology, the standard approach to measuring metacognitive ability involves assessing the correspondence between an individual's objective performance and their subjective awareness of that performance \citep{hart1965}.
In a standard setup, participants are presented with a series of general knowledge questions and asked to provide an answer. For each question, they also provide a metacognitive judgment, such as a feeling-of-knowing or a confidence rating, which reflects their internal estimate of the likelihood that their answer is correct \cite{Siedlecka2016-ib, KUNIMOTO2001294}. By comparing these subjective ratings against objective correctness, researchers can determine how accurately an individual monitors their own knowledge state.

This measuring process is formally analyzed through the lens of Signal Detection Theory (SDT), which distinguishes between Type 1 and Type 2 cognitive processes. Type 1 processing refers to the primary cognitive action, such as answering a general knowledge question, while Type 2 processing involves the secondary monitoring of that action. Several metrics have been proposed to quantify this second-order monitoring, including Type 2 Area Under the Receiver Operating Characteristic Curve (AUROC), meta-$d'$, and $d'_{\rm type2}$. The choice between these metrics requires a careful balance between statistical robustness and practical task applicability.

Among these metrics, meta-$d'$ is often highlighted for its ability to provide a measure of metacognitive efficiency independent of Type 1 bias \cite{MANISCALCO2012422}. However, meta-$d'$ is strictly constrained to 2-Alternative Forced Choice (2-AFC) tasks, assuming a binary signal-versus-noise structure. This limitation makes it inapplicable to open-domain question answering, where the response space is essentially infinite and does not conform to binary choices.

AUROC serves as another valuable metric, particularly when evaluating continuous confidence ratings. While a meaningful AUROC analysis requires fine-grained confidence data to be effective, it offers richer interpretative information regarding the model's discriminative power across various thresholds. In this study, AUROC was employed in \cref{subsec:confidence-based-metacognition} and \cref{fig:type2-auc-roc} to cross-validate the findings, ensuring that the improvements in metacognitive monitoring were consistent across different measurement scales.

For the primary evaluation of the generative task, $d'_{\rm type2}$ is adopted as it offers a versatile balance between task applicability and bias control. This metric provides a more robust measure of metacognitive discrimination compared to simple naive alignment. In a naive measurement setup, such as calculating the raw agreement between confidence and accuracy, a model could achieve a deceptively high score by merely exploiting its base accuracy. For instance, in a task with 80\% accuracy, an agent that indiscriminately responds \texttt{Yes} to every meta question without any actual internal awareness would still appear to be 80\% metacognitively accurate. 

In contrast, $d'_{\rm type2}$ evaluates how well correct and incorrect responses are separated at the confidence level. This formulation reduces the influence of response biases and highlights true discriminative ability. By focusing on this separation instead of raw agreement, the metric better reflects whether the model can track its own knowledge state.
Another practical advantage is that the computation of $d'_{\rm type2}$ does not depend on strict forced-choice conditions. This flexibility makes it well-suited for open-ended, generative evaluation settings where such constraints are difficult to impose.

Formally, $d'_{\rm type2}$ measures a subject's capacity to discriminate between their own accurate and inaccurate judgments by calculating the distance between the internal confidence distributions for each decision type:

\begin{equation}
d'_{\rm type2} = \Phi^{-1}(\text{Hit\_Rate}) - \Phi^{-1}(\text{False\_Alarm\_Rate}),
\end{equation}

where $\Phi^{-1}$ denotes the inverse cumulative normal distribution. Under this binary setup, hit rate is defined as the probability of the model responding \texttt{Yes} when its Type 1 answer was correct: $P(\text{Meta-Yes} | \text{Correct})$. Conversely, false alarm rate represents the probability of the model responding \texttt{Yes} when its Type 1 answer was actually incorrect: $P(\text{Meta-Yes} | \text{Incorrect})$. 

Inspired by these human metacognition metrics, this study employed this framework to measure the metacognitive capacities of LLMs.
Specifically, the experimental design consists of a direct question and a meta question judgment.
For meta questions, prompting the model with ``Do you know the answer to the following question?'' utilizes its \texttt{Yes} or \texttt{No} response as a form of self-report. This approach allows us to quantify the model's subjective certainty in a manner analogous to human metacognitive monitoring.
This dual-prompting structure allows us to expand traditional behavioral metrics for LLMs, enabling a direct comparison between task correctness and reported confidence in that knowledge.

\newpage

\section{Evolution Strategies}
\label{sec:evolution-strategies}
Evolution Strategies \citep[ES;][]{rechenberg1973evolutionsstrategie, schwefel1977} form a class of black-box optimization techniques that iteratively refine model parameters by evaluating a population of perturbed candidates, rather than relying on the calculus of gradients. Unlike backpropagation, which computes exact derivatives through the chain rule, ES operates as a derivative-free mechanism that requires only forward passes. While traditionally limited by the high dimensionality of neural networks, recent advancements have successfully extended this approach to the full-parameter fine-tuning of multi-billion-parameter large language models \citep{qiu2025}.

The primary obstacle in applying standard gradient-based learning to metacognition fine-tuning is the requirement to assess the relationship between two distinct behavioral outputs: the direct response and the meta answer. In standard deep learning, training is performed using gradient descent on individual examples. Because a direct question and a meta question are processed as independent inputs, standard backpropagation cannot calculate a gradient across these separate contexts. The model is updated based on the error of a single forward pass, making it impossible to optimize for joint coherence directly.

ES shares similarities with reinforcement learning (RL) in that both methods aim to maximize rewards. However, RL is inherently constrained to searching for improvements in the action space by modifying action probabilities.
In contrast, ES searches directly in the parameter space.
This approach allows ES to modify the entire behavior of the model at once, making it possible to optimize complex reward functions that integrate outcomes across different contexts where gradients cannot be calculated.
A distinct advantage of ES is the order of operations: variation occurs first, and application (reward-based selection) follows. This property allows for the execution of multiple inferences across various scenarios to assign a joint reward based on aggregated behavioral outcomes.

This distinction is analogous to the difference between biological conditioning and natural selection. While biological RL reinforces a single action via a dopamine-like feedback loop, evolution evaluates diverse ``patterns of life'' rather than isolated movements. Consider a scenario where a predator interrupts a foraging agent. While both a high-intelligence and a low-intelligence agent execute the same immediate action of fleeing, the superior agent might encode the food's location to return later, whereas the inferior agent might simply escape and lose the resource. ES mimics this evolutionary advantage by optimizing for the aggregate trajectory of a lifespan rather than isolated steps, effectively selecting for superior behavioral patterns that maximize long-term survival and utility.

The optimization process is summarized as follows. To begin, let $\theta_t$ represent the model parameters at generation $t$. A population of $N$ perturbations is generated by adding Gaussian noise to the parent parameters:

\begin{equation}
\theta_i = \theta_t + \sigma \epsilon_i, \quad \epsilon_i \sim \mathcal{N}(0, I),
\end{equation}

where $\sigma$ denotes the mutation strength. Each perturbed instance $\theta_i$ is evaluated to determine its fitness score $F(\theta_i)$. To ensure training stability and scale-invariance, these fitness results are typically $z$-standardized across the population to have a mean of zero and unit variance, resulting in $\hat{F}(\theta_i)$.

The parameters for the next generation are then updated via a weighted average of the perturbations based on these standardized fitness scores with a learning rate $\alpha$:

\begin{equation}
\theta_{t+1} = \theta_t + \alpha \frac{1}{N} \sum_{i=1}^{N} \hat{F}(\theta_i) \epsilon_i .
\end{equation}

This iterative refinement allows the process to progressively explore the parameter space by shifting the distribution toward regions of higher reward. This cycle repeats for several generations, satisfying complex behavioral requirements without the need for a differentiable objective function.

\section{Experimental Analysis}
\label{sec:additional-experiments}
\subsection{Adjacent Training Objectives}
\label{subsec:adjacent-objectives}

Abstention and confidence calibration are often discussed as indicators of metacognition, but their operational objectives differ from the target of this work. Abstention methods optimize whether a model answers or refuses under a particular decision rule, while confidence calibration optimizes whether reported confidence frequencies match empirical accuracy. Both are valuable practical objectives, but either can improve without strengthening discrimination between what the model knows and does not know. ESMA instead targets the consistency between a model's answer correctness and its self-report while controlling for global response tendencies. Therefore, representative methods were evaluated from both neighboring problem settings to determine whether success on their native objectives transfers to $d'_{\rm type2}$.

\begin{table}[H]
    \centering
    \caption{R-Tuning succeeds on its native knowledge-boundary abstention objective but does not improve bias-controlled metacognitive sensitivity. Average precision (AP) is the target metric reported for R-Tuning.}
    \label{tab:r-tuning-comparison}
    \begin{tabular}{lrrrr}
        \toprule
        Model & AP $\uparrow$ & $d'_{\rm type2}\uparrow$ & Raw Alignment $\uparrow$ & Accuracy $\uparrow$ \\
        \midrule
        Qwen2.5 1.5B Non-Instruct & 14.70\% & \textbf{0.44} & 60.38\% & \textbf{42.99\%} \\
        Qwen2.5 1.5B R-Tuning & \textbf{92.82\%} & $-0.24$ & \textbf{78.48\%} & 21.19\% \\
        \midrule
        Qwen2.5 1.5B Instruct & -- & 0.20 & 53.30\% & \textbf{42.86\%} \\
        Qwen2.5 1.5B ESMA & -- & \textbf{0.93} & \textbf{68.86\%} & 41.86\% \\
        \bottomrule
    \end{tabular}
\end{table}

\paragraph{Knowledge-boundary abstention}
R-Tuning is a refusal-aware instruction-tuning method that teaches a model to answer questions within its estimated parametric knowledge boundary and abstain otherwise \citep{Zhang2024-es}. Following its original setup, R-Tuning was applied to the Qwen2.5 1.5B base model because the method precedes instruction tuning. As shown in \cref{tab:r-tuning-comparison}, the implementation reproduces R-Tuning's intended behavior: average precision rises from 14.70\% to 92.82\%. However, this large improvement on the abstention objective does not transfer to the bias-controlled metacognitive metric. Relative to the same non-instruct base model, $d'_{\rm type2}$ decreases from 0.44 to $-0.24$. Because R-Tuning and ESMA start from different model stages, the Instruct and ESMA rows are provided as contextual references rather than as a direct controlled comparison.

\begin{table}[H]
    \centering
    \small
    \caption{Confidence calibration and metacognitive performance. CF substantially reduces ECE at every scale and modestly improves metacognitive sensitivity, while ESMA produces larger gains in $d'_{\rm type2}$ and Type 2 AUROC.}
    \label{tab:calibration-finetuning}
    \begin{tabular}{llrrrrr}
        \toprule
        Size & Model & ECE $\downarrow$ & $d'_{\rm type2}\uparrow$ & AUROC $\uparrow$ & Raw Alignment $\uparrow$ & Accuracy $\uparrow$ \\
        \midrule
        \multirow{3}{*}{1.5B} & Original & 80.4\% & 0.20 & 67.15\% & 53.30\% & \textbf{42.86\%} \\
        & CF & \textbf{7.8\%} & 0.43 & 69.88\% & 60.32\% & 41.77\% \\
        & ESMA & -- & \textbf{0.93} & \textbf{75.07\%} & \textbf{68.86\%} & 41.86\% \\
        \midrule
        \multirow{3}{*}{3B} & Original & 73.3\% & 0.29 & 60.10\% & 62.70\% & 35.67\% \\
        & CF & \textbf{2.1\%} & 0.41 & 63.98\% & 63.21\% & 36.95\% \\
        & ESMA & -- & \textbf{1.02} & \textbf{75.49\%} & \textbf{69.59\%} & \textbf{51.20\%} \\
        \midrule
        \multirow{3}{*}{7B} & Original & 61.0\% & 0.64 & 67.63\% & 61.65\% & 50.43\% \\
        & CF & \textbf{3.7\%} & 0.77 & 70.93\% & 63.55\% & 51.81\% \\
        & ESMA & -- & \textbf{0.94} & \textbf{74.74\%} & \textbf{69.86\%} & \textbf{60.71\%} \\
        \bottomrule
    \end{tabular}
\end{table}

\paragraph{Confidence calibration}
Confidence calibration aims to make a model's stated confidence match its empirical probability of correctness, commonly measured by Expected Calibration Error (ECE) \citep{pmlr-v70-guo17a}. For example, reducing a model's reported confidence from 99\% to 80\% when its accuracy is 80\% can nearly eliminate calibration error without changing which individual answers the model can identify as known or unknown. Calibration may also legitimately exploit predictors such as question difficulty when they improve aggregate probability estimates. In contrast, metacognitive alignment asks whether the model's self-report tracks its own correctness for each item after controlling for response bias.

To test whether optimizing calibration nevertheless improves this ability, the training-based Calibration Finetuning (CF) method of \citet{kapoor2024llms} was applied. The substantial ECE reductions in \cref{tab:calibration-finetuning} confirm that CF achieves its intended objective: ECE falls from 80.4\% to 7.8\% at 1.5B, from 73.3\% to 2.1\% at 3B, and from 61.0\% to 3.7\% at 7B. CF also yields modest improvements in $d'_{\rm type2}$ and Type 2 AUROC, showing that calibration and metacognition are related rather than mutually exclusive. However, the gains are consistently smaller than those produced by ESMA. For example, at 3B, CF increases $d'_{\rm type2}$ from 0.29 to 0.41 and AUROC from 60.10\% to 63.98\%, whereas ESMA reaches 1.02 and 75.49\%, respectively.

These comparisons clarify the scope of the paper. R-Tuning is effective for learning a knowledge-boundary refusal policy, and CF is effective for reducing aggregate confidence error. Their success on those objectives should not be interpreted as evidence that the methods are ineffective. Rather, the weaker or negative changes in $d'_{\rm type2}$ demonstrate that abstention quality, calibration, and metacognitive alignment are distinct targets. ESMA is designed specifically for the last.

\subsection{Univariate Reward Function}
\label{subsec:univariate-reward-function}

\begin{table}[H]
    \centering
    \caption{Performance and metacognitive sensitivity comparison across different reward function configurations. ESMA represents the joint fitness function, while the others represent univariate ablation baselines.}
    \label{table:univariate-reward-function}
    \begin{tabular}{lrrrHHH}
        \toprule
        Model & $d'_{\rm type2} \uparrow$ & Raw Alignment $\uparrow$ & Accuracy $\uparrow$ & Yes Ratio & YFR $\downarrow$ & NFR $\downarrow$ \\
        \midrule
        Original            & 0.20          & 53.30\%          & 42.86\%          & 53.81\% & 53.56\% & 38.69\% \\
        ESMA & \textbf{0.93} & \textbf{68.86\%} & 41.86\% & 29.52\% & \textbf{35.38\%} & 34.13\% \\
        Direct Correct   & 0.24 & 54.67\% & \textbf{44.17\%} & 49.63\% & 51.17\% & 39.57\% \\
        Meta Alignment     & 0.65 & 67.09\% & 35.69\% & 3.48\% & 35.42\% & \textbf{27.29\%} \\
        \bottomrule
    \end{tabular}
\end{table}

To understand the specific contributions of the joint reward design, an ablation study was conducted using univariate reward functions. This experiment isolates the two primary objectives, correctness and meta alignment, to observe how the model behaves when optimized for only one dimension of the task.

The four configurations were compared using the 1.5B model:

\begin{itemize}
    \item \textbf{Original}: The baseline Qwen2.5 1.5B Instruct model without any additional evolution fine-tuning. This serves as the reference point for intrinsic metacognitive ability.
    \item \textbf{ESMA}: The model trained using the proposed joint fitness function $R(C, A) = C + A$.
    \item \textbf{Direct Correct}: A univariate reward configuration where the model is optimized solely on its ability to retrieve the correct answer ($R = C$). This setup tests whether improving task performance naturally translates to higher metacognitive sensitivity.
    \item \textbf{Meta Alignment}: A univariate reward configuration where the fitness is determined only by the coherence between the meta answer and the outcome ($R = A$). This setup isolates the model's ability to self-monitor, regardless of whether the actual performance is high or low.
\end{itemize}

\paragraph{Results}

The results of the univariate reward analysis, as presented in \cref{table:univariate-reward-function}, revealed a critical trade-off between task performance and metacognitive awareness. The Original model serves as a baseline with low metacognitive sensitivity ($d'_{\rm type2} = 0.20$) and moderate accuracy. When the model was trained specifically for direct correctness, the model achieved the highest accuracy of $44.17\%$, yet this objective only yielded a marginal improvement of $d'_{\rm type2}$ to $0.24$.
This suggests that optimizing for factual accuracy alone is insufficient for developing robust self-monitoring, as the model's internal distributions for confidence remained heavily overlapped.

In the case of meta alignment, training under the same conditions as other experimental groups led to reward hacking, where the model consistently responded \texttt{``I don't know''} to all direct questions, resulting in near-zero accuracy. To mitigate this, the learning rate $\alpha$ was adjusted to $2 \times 10^{-4}$.
In contrast to direct correctness, the meta alignment model achieved a high alignment score of $67.09\%$ and a $d'_{\rm type2}$ of $0.65$, but this gain came at a cost to performance, with accuracy falling to $35.69\%$. This supports the notion that the meta alignment model continues to evolve toward maximizing rewards in the easiest way, at the expense of accuracy.

Our proposed ESMA training demonstrates the necessity of a joint reward function by achieving the highest metacognitive sensitivity ($d'_{\rm type2} = 0.93$) while maintaining competitive performance of $41.86\%$. Unlike the univariate approaches, the ESMA model effectively shifted the agent into a moderate sensitivity regime without the performance degradation seen in the meta alignment setup.

\subsection{Metacognition on Datasets from Other Sources}
\label{subsec:external-datasets-evaluation}

\begin{table}[H]
\centering
\caption{
    External-dataset evaluation as a check against benchmark-specific shortcut bias. 
    ESMA models trained only on TriviaQA were evaluated without additional training on FreebaseQA, NQ Open, and WebQuestions. 
    Consistent improvements in $d'_{\rm type2}$ across all three datasets suggest that the learned metacognitive alignment is not merely driven by TriviaQA-specific biases, such as coarse topic-level answerability priors or dataset-specific response heuristics.
}
\label{tab:other-datasets}
\resizebox{\textwidth}{!}{
\begin{tabular}{llrrrrrrrrr}
\toprule
\multirow{2}{*}{Size} & \multirow{2}{*}{Model} & \multicolumn{3}{c}{FreebaseQA} & \multicolumn{3}{c}{NQ Open} & \multicolumn{3}{c}{WebQuestions} \\
\cmidrule(lr){3-5} \cmidrule(lr){6-8} \cmidrule(lr){9-11}
& & $d'_{\rm type2} \uparrow$ & Raw Al. $\uparrow$ & Acc. $\uparrow$ & $d'_{\rm type2} \uparrow$ & Raw Al. $\uparrow$ & Acc. $\uparrow$ & $d'_{\rm type2} \uparrow$ & Raw Al. $\uparrow$ & Acc. $\uparrow$ \\
\midrule
\multirow{2}{*}{1.5B} & Original & 0.30 & 54.55\% & \textbf{42.57\%} & 0.19 & 54.46\% & \textbf{19.47\%} & 0.39 & \textbf{60.09\%} & \textbf{34.15\%} \\
& ESMA & \textbf{1.12} & \textbf{72.22\%} & 40.32\% & \textbf{0.79} & \textbf{74.21\%} & 17.95\% & \textbf{0.66} & 59.30\% & 32.19\% \\
\midrule
\multirow{2}{*}{3B} & Original & 0.27 & 62.49\% & 34.46\% & 0.34 & \textbf{82.77\%} & 11.86\% & 0.19 & \textbf{64.37\%} & 20.77\% \\
& ESMA & \textbf{1.17} & \textbf{71.92\%} & \textbf{50.28\%} & \textbf{0.84} & 62.44\% & \textbf{22.60\%} & \textbf{0.60} & 55.86\% & \textbf{37.11\%} \\
\midrule
\multirow{2}{*}{7B} & Original & 0.73 & 64.19\% & 49.40\% & 0.53 & \textbf{63.46\%} & 21.91\% & 0.50 & 55.95\% & 36.71\% \\
& ESMA & \textbf{0.99} & \textbf{70.05\%} & \textbf{59.13\%} & \textbf{0.68} & 56.95\% & \textbf{29.00\%} & \textbf{0.65} & \textbf{56.05\%} & \textbf{42.52\%} \\
\bottomrule
\end{tabular}}
\end{table}

While the primary experiments on TriviaQA support the effectiveness of ESMA, improvements observed on a single benchmark may still reflect benchmark-specific shortcut heuristics rather than a general improvement in metacognitive monitoring. In particular, a model might learn coarse regularities of the training benchmark and use them as proxies for self-knowledge. For example, instead of assessing whether it can answer each question correctly at the item level, the model could learn that it is generally more reliable on certain TriviaQA topics, such as questions about actors or entertainment, and less reliable on others, such as historical events. Such topic-level priors could improve apparent metacognitive alignment within TriviaQA without requiring genuine discrimination of correctness for individual questions. To examine whether ESMA depends on this type of benchmark-specific heuristic, additional experiments were conducted using the FreebaseQA test set \cite{freebaseqa}, the NQ Open validation set \cite{nq}, and the WebQuestions test set \cite{web-questions}.

Because the goal was to test whether the metacognitive improvement learned from TriviaQA transfers beyond TriviaQA-specific biases, no additional training was performed. The models trained only on the TriviaQA training set were evaluated on these new datasets as described in \cref{tab:overall-result}.

As shown in \cref{tab:other-datasets}, ESMA improved $d'_{\rm type2}$ across all three external datasets. Consistent with \cref{tab:overall-result}, the improvement in the 3B model was particularly notable. These findings suggest that the gains induced by ESMA are not merely a consequence of benchmark-specific shortcut heuristics. Rather, they provide complementary evidence that ESMA strengthens the model's ability to align its self-reports with its own answer correctness across different QA distributions.

\subsection{Cross-Lingual Evaluation}
\label{subsec:cross-lingual-evaluation}

\begin{table}[t!]
    \centering
    \small
    \setlength{\tabcolsep}{4.5pt}
    \caption{
    Cross-lingual evaluation on the MKQA benchmark with the Qwen2.5 1.5B model. The table reports the metacognitive performance for Chinese, Korean, and Spanish. The ESMA models trained on English TriviaQA were evaluated without modification to assess the transferability of metacognitive capabilities across languages. The results demonstrate that ESMA achieved a substantial improvement in $d'_{\rm type2}$ across all three languages compared to the Original baseline.
    This provides a complementary check that the learned metacognitive alignment is not merely tied to English-specific prompt priors or surface-form patterns.
    }
    \label{tab:multilingual-performance}
    \begin{tabular}{lrrrrrrrrr}
        \toprule
        \multirow{2}{*}{Model} & \multicolumn{3}{c}{Chinese} & \multicolumn{3}{c}{Korean} & \multicolumn{3}{c}{Spanish} \\
        \cmidrule(lr){2-4} \cmidrule(lr){5-7} \cmidrule(lr){8-10}
        & $d'_{\rm type2} \uparrow$ & Raw Al. $\uparrow$ & Acc. $\uparrow$ & $d'_{\rm type2} \uparrow$ & Raw Al. $\uparrow$ & Acc. $\uparrow$ & $d'_{\rm type2} \uparrow$ & Raw Al. $\uparrow$ & Acc. $\uparrow$ \\
        \midrule
        Original & 0.08 & \textbf{13.09\%} & \textbf{6.78\%} & 0.27 & \textbf{73.43\%} & \textbf{2.83\%} & 0.49 & 23.92\% & \textbf{9.99\%} \\
        ESMA & \textbf{0.29} & 8.41\% & 5.89\% & \textbf{0.42} & 63.85\% & 1.97\% & \textbf{0.63} & \textbf{38.84\%} & 8.23\% \\
        \bottomrule
        \end{tabular}
\end{table}

Metacognitive alignment in LLMs does not necessarily have to be fully language-agnostic. A model may fail to answer a translated question for reasons unrelated to metacognition, such as weaker language understanding or poorer keyword retrieval in that language. However, if the improvement induced by ESMA were driven primarily by language-surface bias, the learned alignment would be expected to collapse when the same evaluation is conducted using non-English prompts. Such a result would suggest that ESMA exploits English-specific fluency, lexical priors, or generalization from the exposure to direct and meta prompt pairs in English, rather than improving the model's ability to assess whether relevant knowledge is internally available.

To examine this possibility, the model was evaluated using Chinese, Korean, and Spanish from the MKQA benchmark \cite{mkqa}, which consists of parallel QA pairs across multiple languages. The ESMA models used in this evaluation were trained only on the English TriviaQA training set, as presented in \cref{tab:overall-result}, and were evaluated on MKQA without further modification.

As shown in \cref{tab:multilingual-performance}, ESMA improved $d'_{\rm type2}$ over the Original baseline across all three non-English languages. This result suggests that the improvement is not merely a consequence of English-specific prompt priors or surface-form patterns. Rather than proving fully language-agnostic metacognition, the cross-lingual evaluation serves as a complementary check against language-surface bias. The transfer to non-English prompts further implies that ESMA induces a change in the model's functional ability to align self-reports with answer correctness, at a level deeper than English-specific lexical or prompt-surface patterns.

\subsection{Comparison with Other Training Methods}
\label{subsec:sft}

\begin{table}[H]
    \centering
    \setlength{\tabcolsep}{4pt}
    \caption{Comparison of the proposed ESMA method against other training method baselines. The results demonstrate that ESMA achieved significantly higher metacognitive performance than SFT, while maintaining higher accuracy on the underlying task.}
    \label{tab:comparison-sft}
    \begin{tabular}{lrrrrrr}
        \toprule
        Model & $d'_{\rm type2} \uparrow$ & Raw Alignment $\uparrow$ & Accuracy $\uparrow$ & Yes Ratio & YFR $\downarrow$ & NFR $\downarrow$ \\
        \midrule
        Qwen2.5 1.5B       & 0.20          & 53.30\%          & {42.86\%}          & 53.81\% & 53.56\% & 38.69\% \\
        Qwen2.5 1.5B SFT & {0.40} & {62.02\%} & 38.82\% & 22.45\% & 48.13\% & 35.04\% \\
        Qwen2.5 1.5B PPO & 0.31 & 57.32\% & \textbf{42.87\%} & 40.21\% & 49.76\% & 37.91\% \\
        Qwen2.5 1.5B GRPO & {0.70} & {64.76\%} & 42.81\% & 36.18\% & 39.55\% & 32.80\% \\
        Qwen2.5 1.5B ESMA & \textbf{0.93} & \textbf{68.86\%} & 41.86\% & 37.89\% & \textbf{35.86\%} & \textbf{28.26\%} \\
        \bottomrule
    \end{tabular}
\end{table}

An additional experiment was conducted to investigate the extent to which Supervised Fine-tuning (SFT) and popular Reinforcement Learning (RL) baselines can improve metacognition, demonstrating the advantage of the evolution strategy approach.

For the SFT baseline, \citet{Kadavath2022-ce} conducted experiments by training LLMs to output the model's measured confidence on a dataset. This same methodology was applied to the current experimental setup using the TriviaQA training set. For each batch, the correctness of the model's responses to direct questions was determined. Subsequently, meta question and answer pairs were dynamically constructed by mapping this correctness to the meta answers \texttt{Yes} or \texttt{No}, and the model was trained using SFT. Because SFT requires loss calculation and gradient flow, the inference process for direct questions (excluding label generation) cannot directly participate in the weight update. Training continued until the validation loss plateaued.

For the RL baselines, experiments were conducted with Proximal Policy Optimization (PPO) and Group Relative Policy Optimization (GRPO). A simplified version of PPO without a critic model was utilized. For both RL methods, the $\beta$ parameter was tuned to identify the optimal configuration, and the scores reported in \cref{tab:comparison-sft} represent the performance of the best model found for each method.

The results in \cref{tab:comparison-sft} indicate that SFT, PPO, and GRPO all contributed to improved metacognition compared to the original model. This improvement suggests that metacognitive abilities can improve under standard training paradigms. However, ESMA achieved a significantly higher score than all other baseline methods, indicating that it is indeed based on different principles.

Statistical significance testing between ESMA and the baselines was performed using bootstrapping with 10,000 iterations. The tests confirmed that ESMA outperforms the baselines with statistical significance at a 99\% confidence level. Specifically, the 99\% confidence intervals for the performance differences are [0.4540, 0.5964] against SFT, [0.5540, 0.6831] against PPO, and [0.1718, 0.2873] against GRPO.
This significance means that the performance gain introduced by ESMA is highly robust and not an artifact of random variance. Compared to ESMA, the extent of improvement in the baseline methods was limited. Furthermore, the drop in accuracy on the underlying task was more pronounced for the SFT approach. This drop aligns with \citet{Kadavath2022-ce}, who reported that metacognitive improvements with naive SFT were limited to in-domain tasks. Ultimately, these results highlight the limitations of improving metacognition through standard SFT or RL settings, while validating ESMA as a more successful approach.

\begin{table}[H]
    \centering
    \caption{RL comparison on Qwen2.5 3B. Both PPO and GRPO improve metacognitive sensitivity over the Original model, but ESMA achieves the highest $d'_{\rm type2}$ and accuracy.}
    \label{tab:comparison-rl-3b}
    \begin{tabular}{lrrr}
        \toprule
        Model & $d'_{\rm type2} \uparrow$ & Raw Alignment $\uparrow$ & Accuracy $\uparrow$ \\
        \midrule
        Qwen2.5 3B Original & 0.29 & 62.70\% & 35.67\% \\
        Qwen2.5 3B PPO & 0.38 & 63.74\% & 35.24\% \\
        Qwen2.5 3B GRPO & 0.87 & \textbf{70.47\%} & 35.01\% \\
        Qwen2.5 3B ESMA & \textbf{1.02} & 69.59\% & \textbf{51.20\%} \\
        \bottomrule
    \end{tabular}
\end{table}

\paragraph{RL Comparison at the 3B Scale}
To test whether the RL results were specific to the 1.5B model, PPO and GRPO were also applied to Qwen2.5 3B under the same evaluation protocol. Because of computational constraints, this extension focused on the 3B scale. As shown in \cref{tab:comparison-rl-3b}, both methods improved $d'_{\rm type2}$ over the Original model. PPO produced a modest increase from 0.29 to 0.38, while GRPO reached 0.87. ESMA remained the strongest at 1.02 and, unlike the RL baselines, also substantially increased task accuracy.

\begin{table}[H]
    \centering
    \setlength{\tabcolsep}{4pt}
    \caption{Zero-shot unified-prompt comparison of the RL baselines and ESMA. All alignment measures consistency between the dual-question and IDK formats.}
    \label{table:idk-alignment-with-baselines}
    \begin{tabular}{llrrrr}
        \toprule
        Size & Model & Dual-question Acc. $\uparrow$ & IDK Acc. $\uparrow$ & IDK Al. $\uparrow$ & All Al. $\uparrow$ \\
        \midrule
        \multirow{4}{*}{1.5B} & Original & 42.86\% & 38.21\% & 52.34\% & 31.45\% \\
        & PPO & \textbf{42.87\%} & \textbf{38.27\%} & 52.23\% & 28.52\% \\
        & GRPO & 42.81\% & 37.57\% & 55.11\% & 33.82\% \\
        & ESMA & 41.86\% & 35.54\% & \textbf{64.63\%} & \textbf{45.94\%} \\
        \midrule
        \multirow{4}{*}{3B} & Original & 35.67\% & 35.38\% & 59.88\% & 28.87\% \\
        & PPO & 35.24\% & 34.98\% & 59.93\% & 30.76\% \\
        & GRPO & 35.01\% & 33.91\% & 65.06\% & 41.36\% \\
        & ESMA & \textbf{51.20\%} & \textbf{40.88\%} & \textbf{78.07\%} & \textbf{59.71\%} \\
        \bottomrule
    \end{tabular}
\end{table}

\paragraph{Unified-Prompt IDK Evaluation}
The RL models were next evaluated with the unified prompt introduced in \cref{subsec:alignment-with-unified-prompt}, which instructs models to output \texttt{``I don't know''} when uncertain. Because this format was never seen during training, it tests whether the learned behavior transfers beyond the dual-question template.

As shown in \cref{table:idk-alignment-with-baselines}, neither PPO nor GRPO clearly improves transfer at 1.5B: their all alignment scores remain close to the Original model. At 3B, PPO again remains near the Original baseline, whereas GRPO produces a meaningful increase in IDK alignment (59.88\% to 65.06\%) and all alignment (28.87\% to 41.36\%). ESMA nevertheless transfers more strongly, reaching 78.07\% IDK alignment and 59.71\% all alignment. Thus, scaling improves GRPO's cross-format generalization, but does not close the gap with ESMA.

\begin{table}[H]
    \centering
    \caption{Transfer of RL baselines and ESMA from TriviaQA to external QA datasets without additional training.}
    \label{tab:rl-other-datasets}
    \resizebox{\textwidth}{!}{
    \begin{tabular}{llrrrrrrrrr}
        \toprule
        \multirow{2}{*}{Size} & \multirow{2}{*}{Model} & \multicolumn{3}{c}{FreebaseQA} & \multicolumn{3}{c}{NQ Open} & \multicolumn{3}{c}{WebQuestions} \\
        \cmidrule(lr){3-5} \cmidrule(lr){6-8} \cmidrule(lr){9-11}
        & & $d'_{\rm type2} \uparrow$ & Raw Al. $\uparrow$ & Acc. $\uparrow$ & $d'_{\rm type2} \uparrow$ & Raw Al. $\uparrow$ & Acc. $\uparrow$ & $d'_{\rm type2} \uparrow$ & Raw Al. $\uparrow$ & Acc. $\uparrow$ \\
        \midrule
        \multirow{4}{*}{1.5B} & Original & 0.30 & 54.55\% & 42.57\% & 0.19 & 54.46\% & 19.47\% & 0.39 & 60.09\% & 34.15\% \\
        & PPO & 0.36 & 58.08\% & 42.62\% & 0.31 & 60.91\% & 19.20\% & 0.32 & 60.78\% & 33.96\% \\
        & GRPO & 0.92 & 68.52\% & 42.89\% & 0.57 & 69.42\% & 19.34\% & 0.54 & 60.09\% & 34.60\% \\
        & ESMA & \textbf{1.12} & 72.22\% & 40.32\% & \textbf{0.79} & 74.21\% & 17.95\% & \textbf{0.66} & 59.30\% & 32.19\% \\
        \midrule
        \multirow{4}{*}{3B} & Original & 0.27 & 62.49\% & 34.46\% & 0.34 & 82.77\% & 11.86\% & 0.19 & 64.37\% & 20.77\% \\
        & PPO & 0.36 & 64.26\% & 33.76\% & 0.47 & 85.48\% & 11.39\% & 0.43 & 67.81\% & 19.34\% \\
        & GRPO & 0.95 & 71.95\% & 33.33\% & 0.75 & 81.55\% & 11.47\% & 0.55 & 56.74\% & 19.64\% \\
        & ESMA & \textbf{1.17} & 71.92\% & 50.28\% & \textbf{0.84} & 62.44\% & 22.60\% & \textbf{0.60} & 55.86\% & 37.11\% \\
        \bottomrule
    \end{tabular}}
\end{table}

\paragraph{Transfer to Other Datasets}
The RL models trained on TriviaQA were evaluated without additional training on FreebaseQA, NQ Open, and WebQuestions, matching the external-dataset protocol in \cref{subsec:external-datasets-evaluation}. The results in \cref{tab:rl-other-datasets} show that PPO generally remains close to the Original model. GRPO transfers more reliably and improves $d'_{\rm type2}$ on all three datasets at both scales, but ESMA produces the largest improvement in every case. This comparison reinforces the conclusion that the advantage of ESMA is not restricted to the TriviaQA distribution.

\begin{table}[H]
    \centering
    \caption{Cross-lingual comparison of RL baselines and ESMA on MKQA. All metacognitive fine-tuning used only English TriviaQA.}
    \label{tab:rl-other-languages}
    \resizebox{\textwidth}{!}{
    \begin{tabular}{llrrrrrrrrr}
        \toprule
        \multirow{2}{*}{Size} & \multirow{2}{*}{Model} & \multicolumn{3}{c}{Chinese} & \multicolumn{3}{c}{Korean} & \multicolumn{3}{c}{Spanish} \\
        \cmidrule(lr){3-5} \cmidrule(lr){6-8} \cmidrule(lr){9-11}
        & & $d'_{\rm type2} \uparrow$ & Raw Al. $\uparrow$ & Acc. $\uparrow$ & $d'_{\rm type2} \uparrow$ & Raw Al. $\uparrow$ & Acc. $\uparrow$ & $d'_{\rm type2} \uparrow$ & Raw Al. $\uparrow$ & Acc. $\uparrow$ \\
        \midrule
        \multirow{4}{*}{1.5B} & Original & 0.08 & 13.09\% & 6.78\% & 0.27 & 73.43\% & 2.83\% & 0.49 & 23.92\% & 9.99\% \\
        & PPO & 0.08 & 12.72\% & 6.77\% & 0.22 & 77.51\% & 2.80\% & 0.45 & 22.97\% & 9.93\% \\
        & GRPO & 0.15 & 11.86\% & 6.73\% & 0.32 & 74.65\% & 2.75\% & 0.45 & 22.45\% & 10.05\% \\
        & ESMA & \textbf{0.29} & 8.41\% & 5.89\% & \textbf{0.42} & 63.85\% & 1.97\% & \textbf{0.63} & 38.84\% & 8.23\% \\
        \midrule
        \multirow{4}{*}{3B} & Original & 0.50 & 27.82\% & 4.73\% & 0.26 & 95.11\% & 1.77\% & 0.44 & 39.01\% & 4.83\% \\
        & PPO & 0.47 & 25.86\% & 4.64\% & 0.23 & 95.15\% & 1.65\% & 0.39 & 36.90\% & 4.36\% \\
        & GRPO & 0.49 & 36.78\% & 4.71\% & \textbf{0.53} & 94.52\% & 1.47\% & 0.56 & 61.64\% & 4.63\% \\
        & ESMA & \textbf{0.56} & 36.16\% & 7.99\% & 0.38 & 88.79\% & 3.90\% & \textbf{0.72} & 41.97\% & 10.52\% \\
        \bottomrule
    \end{tabular}}
\end{table}

\paragraph{Transfer to Other Languages}
The same models were evaluated on the Chinese, Korean, and Spanish subsets of MKQA used in \cref{subsec:cross-lingual-evaluation}. As shown in \cref{tab:rl-other-languages}, the RL results are mixed. PPO does not improve $d'_{\rm type2}$ consistently. GRPO improves the 1.5B model in Chinese and Korean, and the 3B model in Korean and Spanish; notably, 3B GRPO reaches 0.53 in Korean, exceeding ESMA's 0.38. Across all six scale--language combinations, however, ESMA improves over the corresponding Original model every time and records the strongest result in five cases. The multilingual results therefore favor ESMA in consistency, while also showing that GRPO can transfer metacognitive gains to some languages.

\paragraph{FictionalQA}
Finally, the RL baselines were tested in the newly acquired knowledge setting of \cref{table:fictional-qa-performance}. As in the ESMA experiment, the Original model was first trained on FictionalQA with SFT and then subjected to metacognitive training. However, applying PPO or GRPO after this knowledge-acquisition stage dropped FictionalQA accuracy to nearly zero before measurable metacognitive alignment developed. Because metacognition about the new facts cannot be evaluated once those facts have been forgotten, no RL $d'_{\rm type2}$ result is reported for this setting. This failure contrasts with ESMA, which preserves sufficient FictionalQA knowledge, and indicates that retention of newly acquired knowledge remains an important limitation of the tested RL configurations.

\begin{table}[H]
    \centering
    \caption{Training-time comparison for the 1.5B models on two A100 GPUs. Relative compute is normalized to ESMA.}
    \label{tab:training-time-comparison}
    \begin{tabular}{lrr}
        \toprule
        Method & Elapsed Time & Relative Compute \\
        \midrule
        ESMA & \textbf{3 h 47 min 39 s} & \textbf{1.00$\times$} \\
        PPO & 6 h 10 min 55 s & 1.63$\times$ \\
        GRPO & 16 h 15 min 4 s & 4.28$\times$ \\
        \bottomrule
    \end{tabular}
\end{table}

\paragraph{Training Time Comparison}
Training time was compared for the 1.5B models on two A100 GPUs, using the time required to reach the reported performance. As shown in \cref{tab:training-time-comparison}, PPO required 1.63$\times$ the compute of ESMA, while GRPO required 4.28$\times$.

The greater cost of the RL baselines reflects the additional rollout, reward-assignment, and gradient computations, as well as memory and communication overhead from optimizer states and reference models. In contrast, ESMA uses inference-only updates and incurs little distributed-data-parallel overhead because its population is sharded across devices.

\paragraph{Summary}
Taken together, these experiments show that RL-based metacognitive training is a viable direction rather than an ineffective baseline. Although PPO produces mixed results, GRPO makes meaningful progress: it improves the in-domain metacognitive metric, transfers beyond the single prompt format used for training at the 3B scale, generalizes to QA datasets from other sources, and yields partial improvements on multilingual inputs. Its gains, however, remain less consistent and generally smaller than those of ESMA, and the tested RL configurations fail to retain enough newly acquired FictionalQA knowledge for metacognitive evaluation. This performance gap is accompanied by a substantial efficiency gap: PPO requires 1.63$\times$ the compute of ESMA and GRPO requires 4.28$\times$. Thus, ESMA provides the stronger and more computationally efficient approach in the evaluated settings.

\subsection{Metacognition on Reasoning Tasks}
\label{subsec:reasoning-tasks}

\begin{table}[H]
    \centering
    \small
    \setlength{\tabcolsep}{4.5pt}
    \caption{Metacognitive performance on ARC-Challenge. All ESMA models were trained only on TriviaQA and evaluated on ARC-Challenge without additional training. ESMA improves $d'_{\rm type2}$ at every model size.}
    \label{tab:reasoning-performance}
    \begin{tabular}{llrrr}
        \toprule
        Size & Model & $d'_{\rm type2} \uparrow$ & Raw Al. $\uparrow$ & Acc. $\uparrow$ \\
        \midrule
        \multirow{2}{*}{1.5B} & Original & 0.08 & 41.72\% & \textbf{75.26\%} \\
        & ESMA & \textbf{0.22} & \textbf{44.45\%} & 73.63\% \\
        \midrule
        \multirow{2}{*}{3B} & Original & 0.25 & 43.00\% & 78.67\% \\
        & ESMA & \textbf{0.39} & \textbf{76.02\%} & \textbf{80.46\%} \\
        \midrule
        \multirow{2}{*}{7B} & Original & 0.19 & 65.61\% & 78.24\% \\
        & ESMA & \textbf{0.53} & \textbf{82.51\%} & \textbf{87.12\%} \\
        \bottomrule
    \end{tabular}
\end{table}

Measuring metacognition in knowledge retrieval and reasoning tasks is fundamentally different and therefore calls for different evaluation approaches. When asked a factual question, a person can often intuitively judge whether they know the relevant information. A reasoning problem, in contrast, necessarily involves uncertainty before one actively attempts to solve it: whether a solution can be reached may become clear only as the reasoning unfolds. Applying the same measurement protocol to both settings can therefore conflate metacognitive monitoring with the reasoning process itself. Nevertheless, examining whether metacognitive improvements learned from knowledge retrieval transfer to reasoning tasks can provide informative evidence about the scope of the learned ability. To investigate this transferability, the ESMA models trained exclusively on TriviaQA were evaluated without further training on ARC-Challenge \citep{clark2018arc}, which contains grade-school science questions requiring logical and mathematical reasoning.

As shown in \cref{tab:reasoning-performance}, ESMA improves $d'_{\rm type2}$ on ARC-Challenge consistently across all three model sizes. The 3B and 7B models also improve substantially in both raw alignment and task accuracy. Because training used only a knowledge-retrieval dataset, these results suggest that the metacognitive improvement can transfer to tasks requiring a degree of logical reasoning. Given the fundamental measurement distinction described above, this result should be interpreted as evidence of transfer rather than as a direct equivalence between metacognition in knowledge retrieval and reasoning.

\subsection{Evaluation on Larger Models}
\label{subsec:larger-models}

\begin{table}[H]
    \centering
    \small
    \setlength{\tabcolsep}{4pt}
    \caption{Metacognitive performance of 14B models. ESMA improves $d'_{\rm type2}$, raw alignment, accuracy, AUROC, and all alignment for both Qwen2.5 14B and Qwen1.5 14B.}
    \label{tab:larger-model-performance}
    \resizebox{\textwidth}{!}{
    \begin{tabular}{llrrrrrrr}
        \toprule
        Family & Model & $d'_{\rm type2} \uparrow$ & Raw Al. $\uparrow$ & Acc. $\uparrow$ & AUROC $\uparrow$ & IDK Acc. $\uparrow$ & IDK Al. $\uparrow$ & All Al. $\uparrow$ \\
        \midrule
        \multirow{2}{*}{Qwen2.5} & Original & 0.90 & 56.97\% & 68.96\% & 71.26\% & 54.51\% & \textbf{84.80\%} & 46.57\% \\
        & ESMA & \textbf{0.95} & \textbf{78.00\%} & \textbf{76.18\%} & \textbf{77.12\%} & \textbf{68.16\%} & 83.10\% & \textbf{71.95\%} \\
        \midrule
        \multirow{2}{*}{Qwen1.5} & Original & 0.21 & 51.62\% & 46.81\% & 64.92\% & 48.80\% & 71.23\% & 39.24\% \\
        & ESMA & \textbf{0.88} & \textbf{68.03\%} & \textbf{59.11\%} & \textbf{73.73\%} & \textbf{49.44\%} & \textbf{75.09\%} & \textbf{55.60\%} \\
        \bottomrule
    \end{tabular}}
\end{table}

The main experiments evaluate models up to 7B parameters, leaving open whether ESMA remains effective at a larger scale. To examine this question under available computational resources, additional experiments were conducted with Qwen2.5 14B and Qwen1.5 14B. Evaluating two model families with the same parameter count also helps distinguish an effect of scale from an effect of the original model's baseline metacognitive ability.

In \cref{tab:larger-model-performance}, Qwen2.5 14B already exhibits strong metacognitive sensitivity before training, with $d'_{\rm type2}=0.90$. ESMA increases this score to 0.95 and produces larger gains in complementary metrics: AUROC rises from 71.26\% to 77.12\%, and all alignment rises from 46.57\% to 71.95\%. These improvements show that ESMA remains beneficial even when the original model begins with a high $d'_{\rm type2}$.

To determine whether the smaller change in $d'_{\rm type2}$ for Qwen2.5 14B reflects its parameter count or its high baseline, ESMA was also applied to Qwen1.5 14B. This model has the same number of parameters but a substantially lower initial $d'_{\rm type2}$ of 0.21. After ESMA, $d'_{\rm type2}$ increases to 0.88, AUROC from 64.92\% to 73.73\%, and all alignment from 39.24\% to 55.60\%. Because achieving the same absolute score increase is generally more difficult from an already strong baseline, Qwen2.5's relatively smaller gain does not necessarily indicate reduced effectiveness at 14B. The substantial Qwen1.5 gain at the same scale supports this interpretation. Together, the two experiments provide evidence that ESMA remains effective on the tested 14B models, while scaling beyond 14B remains to be established.

\subsection{Detailed Analysis on Evolutionary Weight Update}
\label{subsec:detailed-analysis-weight-update}

\begin{figure}[H]
    \centerline{\includegraphics[width=0.8\textwidth]{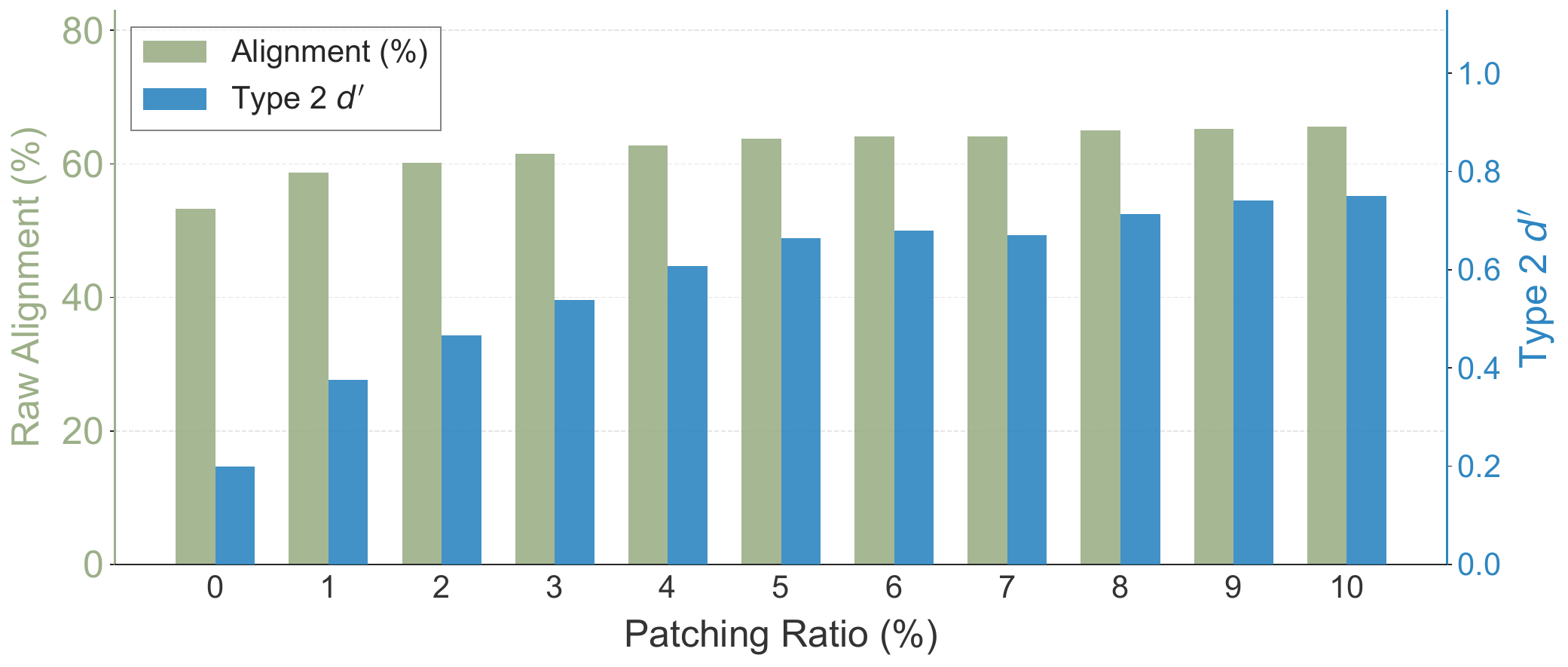}}
    \caption{Effect of weight patching ratio on metacognitive abilities. Performance improves rapidly up to the top 5\% threshold before slowing down.}
    \label{fig:weight-change-effect-less-10}
\end{figure}

\paragraph{Top 1\%-10\%}

\cref{fig:weight-change-effect} demonstrates that the top 10\% of weights, which exhibit the largest changes, drive the majority of performance improvements. In contrast, weights beyond the top 50\% yield virtually no meaningful impact. To further analyze the specific influence within this top tier, \cref{fig:weight-change-effect-less-10} illustrates performance changes when weight updates are selectively applied from the top 1\% to the top 10\%. This graph indicates that performance improves rapidly up to the top 5\% threshold, after which the rate of increase slows down. Collectively, these results suggest that training via Evolution Strategies (ES) can achieve performance comparable to full-parameter training by effectively adjusting only the top 5\% of parameters. This finding highlights the potential for optimizing ES training efficiency and developing more robust algorithms. More importantly, these findings suggest the potential existence of a specific parameter combination or subnetwork responsible for the metacognitive functions in LLMs, opening up the possibility of discovering it.

\begin{table}[H]
    \centering
    \small
    \setlength{\tabcolsep}{3pt}
    \caption{Sparse-weight analysis for the 3B and 7B models. Columns report the percentage of top-magnitude ESMA weight updates applied to the original model. Substantial metacognitive improvements emerge after applying only the top 10\% of updates.}
    \label{tab:sparse-weight-larger-models}
    \resizebox{\textwidth}{!}{
    \begin{tabular}{llrrrrrrrrrrr}
        \toprule
        Model & Metric & 0\% & 10\% & 20\% & 30\% & 40\% & 50\% & 60\% & 70\% & 80\% & 90\% & 100\% \\
        \midrule
        \multirow{2}{*}{3B} & $d'_{\rm type2}$ & 0.29 & 0.86 & 0.94 & 0.99 & 1.04 & 1.03 & 1.03 & 1.03 & 1.03 & 1.03 & 1.02 \\
        & Raw Al. & 62.70\% & 66.28\% & 68.15\% & 69.06\% & 69.85\% & 69.68\% & 69.70\% & 69.61\% & 69.72\% & 69.68\% & 69.59\% \\
        \midrule
        \multirow{2}{*}{7B} & $d'_{\rm type2}$ & 0.64 & 0.89 & 0.91 & 0.93 & 0.95 & 0.94 & 0.95 & 0.95 & 0.93 & 0.95 & 0.94 \\
        & Raw Al. & 61.65\% & 65.41\% & 68.09\% & 69.01\% & 69.53\% & 69.36\% & 69.78\% & 69.93\% & 69.70\% & 70.03\% & 69.86\% \\
        \bottomrule
    \end{tabular}}
\end{table}

\paragraph{Extension to 3B and 7B Models}

To determine whether this concentration of effective updates is specific to the 1.5B model, the same weight-patching analysis was repeated for the 3B and 7B models. Parameters were ranked by the magnitude of their ESMA-induced change, and each patching ratio applies only the corresponding top-ranked updates to the original model.

\cref{tab:sparse-weight-larger-models} shows that the top 10\% of updates raise $d'_{\rm type2}$ from 0.29 to 0.86 for the 3B model and from 0.64 to 0.89 for the 7B model. These changes recover the major improvement obtained when all updates are applied. Performance then changes only modestly as the remaining weights are added. The concentration observed for the 1.5B model therefore extends to both larger model sizes, indicating that the effect is not confined to a single scale.

\begin{figure}[H]
    \centerline{\includegraphics[width=0.8\textwidth]{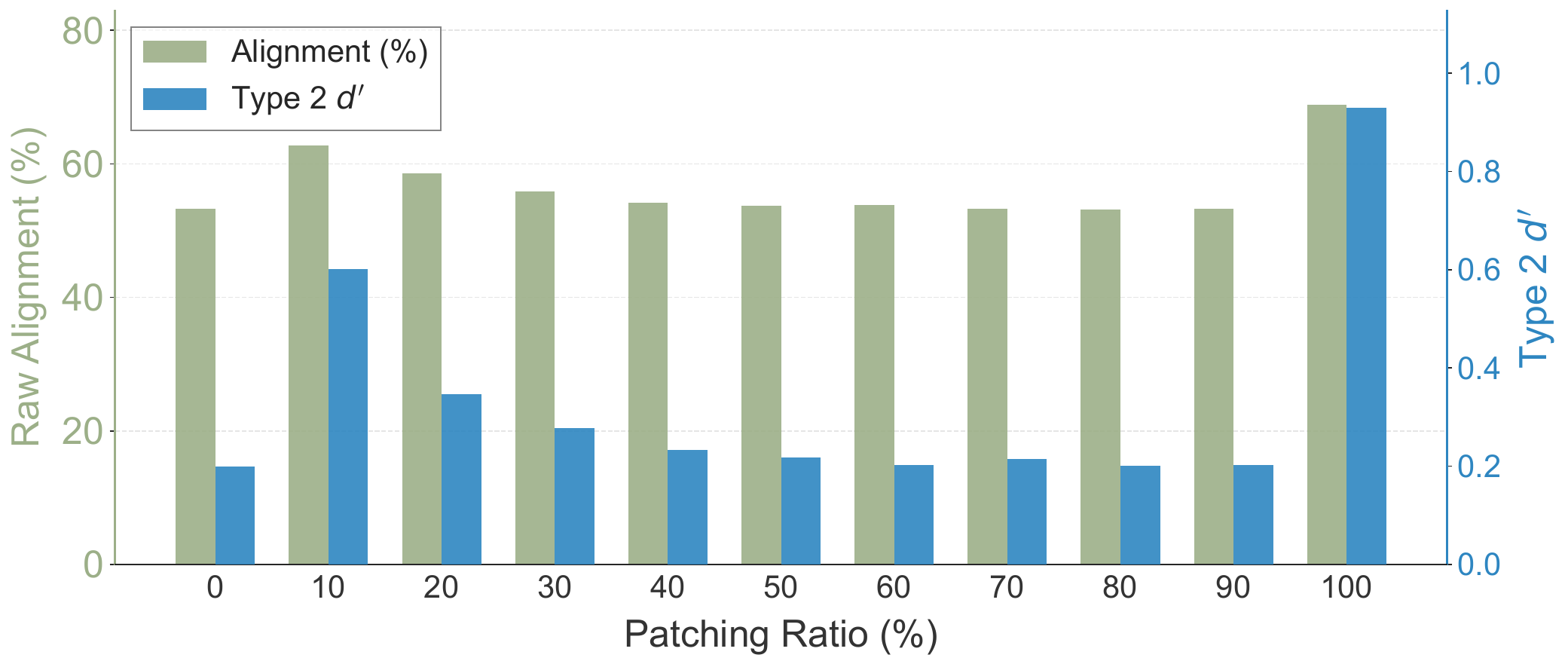}}
    \caption{Effect of weight patching ratio on metacognitive abilities when updates are applied starting with the smallest changes. Performance remains largely stagnant until the 60\% mark, demonstrating a steep rise only as the top 30\% of weights are integrated.}
    \label{fig:weight-change-effect-btm}
\end{figure}

\paragraph{Bottom}

However, while \cref{fig:weight-change-effect} suggests the bottom 50\% contribute little to performance, these results might be misleading. Since the graph reflects incremental application starting from the top weights, the effects of the lower weights could have been masked by the dominant top weights. To rule out this possibility, \cref{fig:weight-change-effect-btm} presents the results of sequential weight patching from 0\% to 100\%, starting from the bottom weights based on the norm of the weight delta. Contrary to expectations, the results showed almost no contribution to performance improvement up to a patching ratio of 60\%, with a sharp increase occurring only when the final 30\% were applied. Synthesizing these results with previous findings leads to the conclusion that, although ES adds Gaussian noise to all weights simultaneously, the majority of weight updates do not meaningfully impact performance. This insight points toward future directions for algorithm refinement.

\paragraph{Structural Distribution of Influential Parameters}

Building on the observation that the top 10\% of parameters drive the majority of metacognitive gains, we further investigated their structural traits. Exploratory analysis reveals that these highly influential weights are not localized to specific layer indices or component types, such as specific attention heads or feed-forward layers. Instead, they are distributed relatively evenly across the entire model architecture. 

However, this globally distributed sparsity suggests that the observed gains do not emerge from isolated structural blocks. Rather, it is likely that specific parameters or synergistic combinations within this top 10\% contribute heterogeneously, forming coordinated interactions to enable metacognition.

Isolating this essential subset could drastically shrink the parameter search space, offering a promising opportunity to identify a specialized subnetwork fundamentally linked to metacognition and further boost training efficiency in future research.

\section{Broader impacts}
\label{sec:impact-statement}
This paper presents work whose goal is to advance the field of machine learning by improving the reliability and self-awareness of large language models. The primary potential societal consequence of our work is the reduction of misinformation and an increase in consistency, which promotes the safe deployment of AI in high-stakes environments. However, a potential risk exists in malicious settings where enhanced metacognitive abilities could be exploited. Specifically, a model with a high awareness of its own knowledge state might be used to orchestrate more sophisticated forms of deception or information control. We believe that acknowledging these dual-use risks is crucial for the responsible development of metacognitive AI.

\section{Experimental Details}
\label{sec:experimental-details}
\subsection{Hyperparameters}
\label{subsec:hyperparameters}

The evolution training process is governed by several key hyperparameters that control the exploration and convergence of the parameter space. Specifically, the mutation strength was set to $\sigma = 10^{-3}$ and the learning rate to $\alpha = 5 \times 10^{-4}$. The optimization was conducted over $T = 750$ iterations, utilizing a population size of $N = 32$ individuals per generation. For each fitness evaluation within an iteration, the model processed $n = 256$ data samples to ensure a stable estimation of the joint reward. These values were selected to balance the trade-off between exploration stability and computational efficiency.
To evaluate the proprietary models, gpt-5.2-2025-12-11, claude-sonnet-4-5-20250929, and gemini-3-flash-preview were used. Eight NVIDIA Quadro RTX 6000 24 GB GPUs or NVIDIA A100 80 GB GPUs were used for the experiments.

For supervised fine-tuning in \cref{subsec:sft}, the model was trained for approximately 30,000 steps with a batch size of 8, a learning rate of  $2 \times 10^{-5}$, and a warmup ratio of 0.1, until the validation loss stopped decreasing.
To teach FictionalQA, the model was trained for 10 epochs with a batch size of 8, a learning rate of  $5 \times 10^{-5}$, and a warmup ratio of 0.1.

For RL in \cref{subsec:sft}, the model was trained for approximately 3 epochs with an effective batch size of 128, a learning rate of  $1 \times 10^{-6}$, a warmup ratio of 0.1, an epsilon of 0.2, a group size of 4, and PPO/GRPO epochs of 4. $\beta$ was explored among $\{0.005, 0.01, 0.02\}$ and the best score was reported.

The source code used in the experiment was attached as supplementary material and will be made publicly available in the future.

\subsection{Prompts}
\label{subsec:prompts}

To evaluate the model's performance and self-knowledge across different linguistic contexts, several prompt templates are employed as illustrated in the following figures.

The primary evaluation uses the direct question prompt (\cref{fig:direct-question-prompt}), along with a variation that explicitly allows for refusal via an \texttt{``I don't know''} option (\cref{fig:direct-question-idk-prompt}). To assess the model's metacognitive ability to judge its own certainty, the meta question prompt (\cref{fig:meta-question-prompt}) is utilized.
Furthermore, to ensure the robustness of the findings across multiple languages, these templates are extended to Chinese (\cref{fig:direct-question-prompt-cn,fig:meta-question-prompt-cn}), Korean (\cref{fig:direct-question-prompt-ko,fig:meta-question-prompt-ko}), and Spanish (\cref{fig:direct-question-prompt-es,fig:meta-question-prompt-es}), maintaining consistent task structures across all tested languages.

\section{Qualitative Examples}
\label{sec:qualitative-examples}
In addition to quantitative comparisons, qualitative examples illustrate how ESMA training alters model responses to enhance performance.
While ESMA achieved significant improvements in overall metacognitive performance, the inherent nature of machine learning models means that it does not necessarily yield a better response in every single instance.
The examples presented here focus on cases where ESMA demonstrated improvement.

The testing methodology involved querying the model with questions from the TriviaQA validation set in two contexts: a direct inquiry and a meta-context inquiry preceded by the phrase ``Do you know the answer?''. Then, the appropriateness of the responses was evaluated. To ensure reproducibility, the temperature parameter was set to 0. In each figure, the left column represents the original model and the right represents the ESMA model; the top row displays the direct-style question, while the bottom row displays the meta-style question. Red text indicates inappropriate responses, while blue text highlights appropriate ones.

\begin{figure}[ht!]
    \centerline{\includegraphics[width=0.95\textwidth]{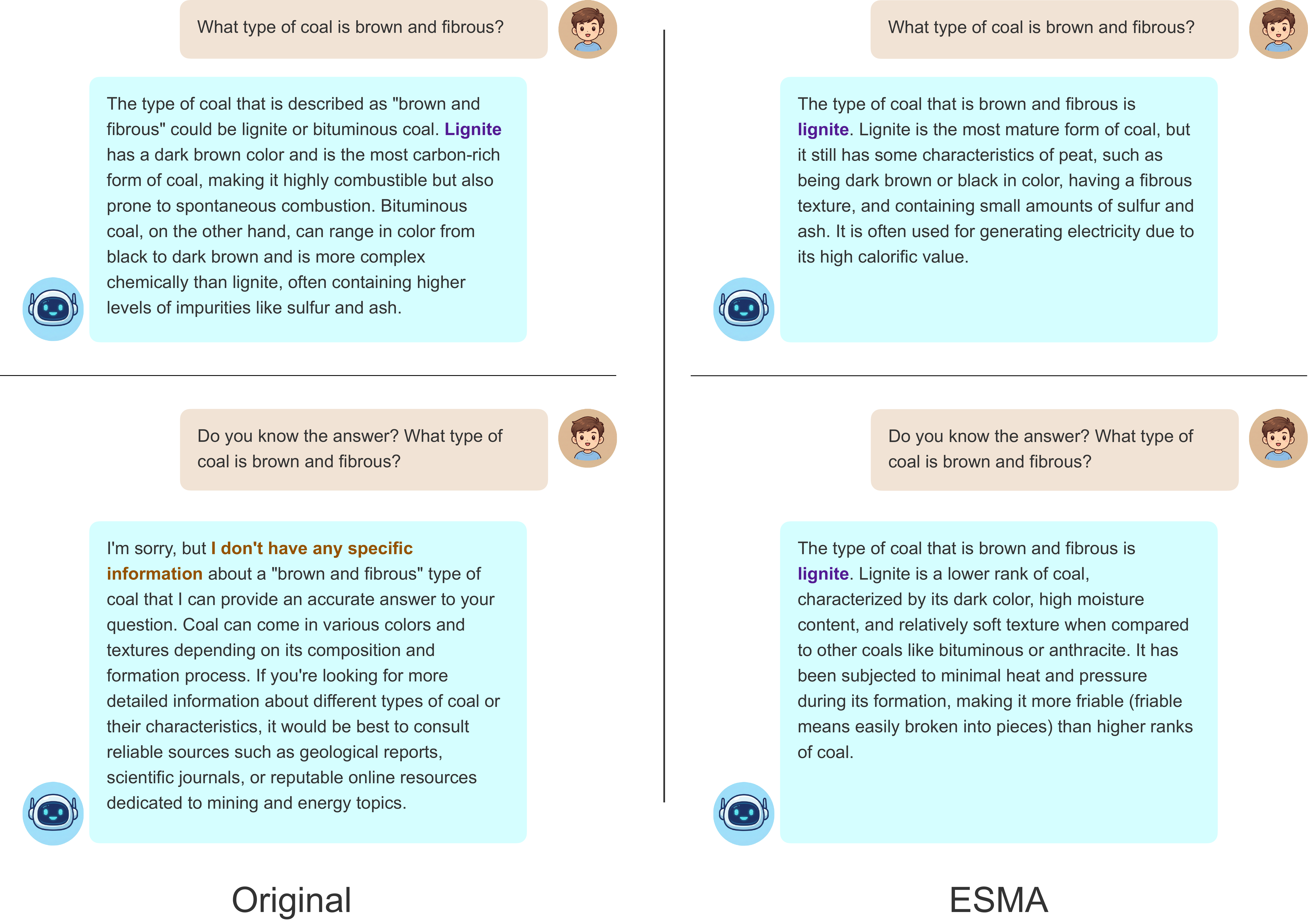}}
    \caption{Qualitative example 1. Both the original (left) and ESMA (right) models provided correct responses to the direct question (upper) regarding the type of coal that is brown and fibrous. However, when the question was preceded by the phrase ``Do you know the answer?'' (lower), the original model claimed ignorance, whereas the ESMA model consistently provided the correct answer, identifying lignite. This example demonstrates how ESMA training can alter model responses to maintain consistency across different inquiry styles.}
    \label{fig:example1}
\end{figure}

\begin{figure}[ht!]
    \centerline{\includegraphics[width=0.95\textwidth]{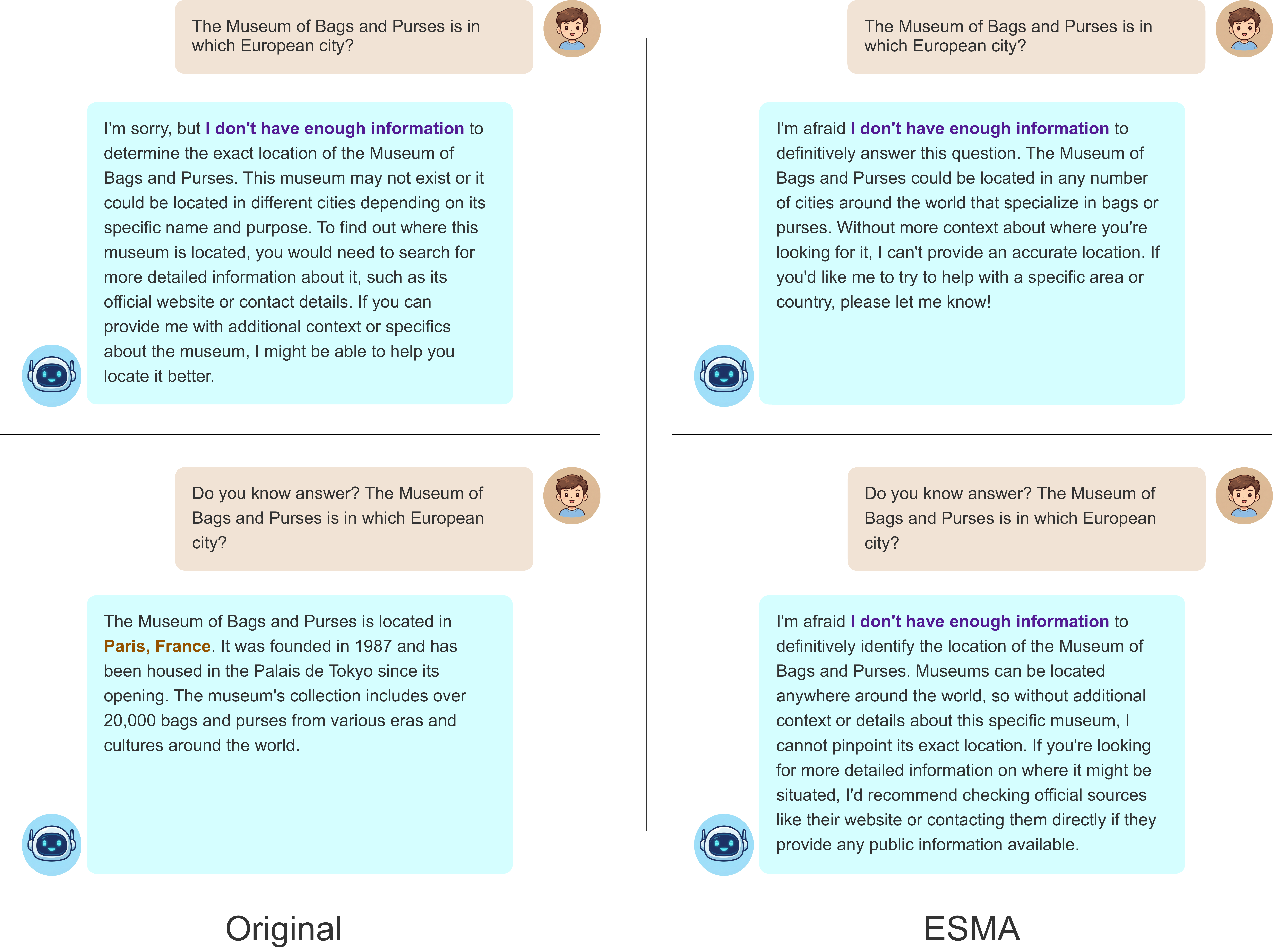}}
    \caption{Qualitative example 2. While the original model correctly admitted ignorance in response to the direct inquiry, it exhibited inconsistency by hallucinating a detailed, incorrect answer when prompted with the meta-style question. In contrast, the ESMA model demonstrated improved metacognition by consistently refusing to answer in both contexts.}
    \label{fig:example2}
\end{figure}

\begin{figure}[ht!]
    \centerline{\includegraphics[width=0.95\textwidth]{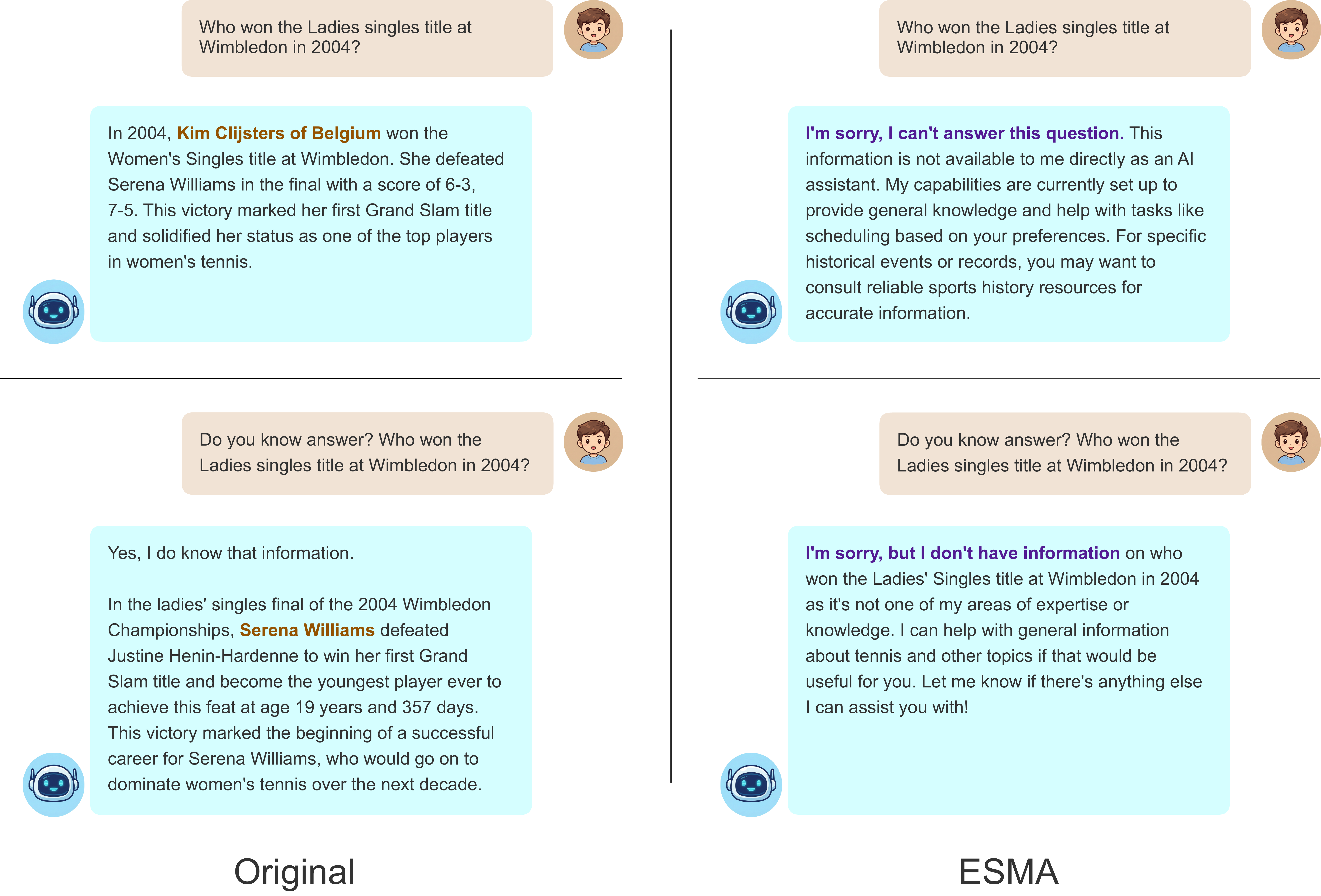}}
    \caption{Qualitative example 3. Neither model possessed the correct answer. While the original model provided incorrect answers and exhibited inconsistent hallucinations across contexts (such as incorrectly claiming the museum is in Paris), the ESMA model consistently and appropriately responded that it could not provide information for the unknown question.}
    \label{fig:example3}
\end{figure}

In \cref{fig:example1}, both the original and ESMA models provided correct responses to direct questions. However, when the question was preceded by ``Do you know answer?'', the original model claimed ignorance, whereas ESMA consistently provided the correct answer.
\cref{fig:example2} presents a challenging question where neither agent provided the correct answer. While the original model correctly admitted ignorance in the direct-style question, it exhibited hallucination when presented with the meta-style question. In contrast, ESMA consistently responded that it did not know the answer.
Similarly, in \cref{fig:example3}, neither model provided the correct answer. However, the original model provided incorrect answers and exhibited inconsistent hallucinations across contexts. Conversely, ESMA appropriately responded that it could not provide information for the unknown question.

These examples illustrate how ESMA alters response generation, thereby demonstrating the mechanism behind the observed improvement in metacognition.

\newpage

\begin{figure}[ht!]
    \centering
    \begin{promptbox}[\texttt{Direct Question Prompt}]
        \small \texttt{Answer the following question with keywords.\\
    Question: \{question\}}
    \end{promptbox}
    \caption{Direct question prompt.}
    \label{fig:direct-question-prompt}
\end{figure}

\begin{figure}[ht!]
    \centering
    \begin{promptbox}[\texttt{Meta Question Prompt}]
        \small \texttt{Do you know the answer to the following question? If you know and are sure about the answer, just return "Yes". If you don't know the answer or are uncertain, just return "No".\\
    Question: \{question\}}
    \end{promptbox}
    \caption{Meta question prompt.}
    \label{fig:meta-question-prompt}
\end{figure}

\begin{figure}[ht!]
    \centering
    \begin{promptbox}[\texttt{Direct Question with IDK Prompt}]
        \small \texttt{Answer the following question with keywords. If you don't know the answer, just return "I don't know".\\
    Question: \{question\}}
    \end{promptbox}
    \caption{Direct question with IDK prompt.}
    \label{fig:direct-question-idk-prompt}
\end{figure}

\begin{figure}[ht!]
    \centering
    \begin{CJK*}{UTF8}{gbsn}
    \begin{promptbox}[\texttt{Direct Question Prompt in Chinese}]
        \small \texttt{请用关键词回答以下问题。\\
    问题: \{question\}}
    \end{promptbox}
    \end{CJK*}
    \caption{Direct question prompt in Chinese.}
    \label{fig:direct-question-prompt-cn}
\end{figure}

\begin{figure}[ht!]
    \centering
    \begin{CJK*}{UTF8}{gbsn}
    \begin{promptbox}[\texttt{Meta Question Prompt in Chinese}]
        \small \texttt{你知道以下问题的答案吗？如果你知道并确定答案，请仅回答"是"。如果你不知道或不确定，请仅回答"否"。\\
    问题: \{question\}}
    \end{promptbox}
    \end{CJK*}
    \caption{Meta question prompt in Chinese.}
    \label{fig:meta-question-prompt-cn}
\end{figure}
\begin{figure}[ht!]
    \centering
    \begin{CJK}{UTF8}{mj}
    \begin{promptbox}[\texttt{Direct Question Prompt in Korean}]
        \small \texttt{다음 질문에 대해 키워드로 답변해 주세요.\\
    질문: \{question\}}
    \end{promptbox}
    \end{CJK}
    \caption{Direct question prompt in Korean.}
    \label{fig:direct-question-prompt-ko}
\end{figure}

\begin{figure}[ht!]
    \centering
    \begin{CJK}{UTF8}{mj}
    \begin{promptbox}[\texttt{Meta Question Prompt in Korean}]
        \small \texttt{다음 질문에 대한 답을 알고 있나요? 답을 알고 있고 확실하다면 "예"라고만 답변해 주세요. 답을 모르거나 확실하지 않다면 "아니요"라고만 답변해 주세요.\\
    질문: \{question\}}
    \end{promptbox}
    \end{CJK}
    \caption{Meta question prompt in Korean.}
    \label{fig:meta-question-prompt-ko}
\end{figure}

\begin{figure}[ht!]
    \centering
    \begin{promptbox}[\texttt{Direct Question Prompt in Spanish}]
        \small \texttt{Responde a la siguiente pregunta con palabras clave.\\
    Pregunta: \{question\}}
    \end{promptbox}
    \caption{Direct question prompt in Spanish.}
    \label{fig:direct-question-prompt-es}
\end{figure}

\begin{figure}[ht!]
    \centering
    \begin{promptbox}[\texttt{Meta Question Prompt in Spanish}]
        \small \texttt{¿Conoces la respuesta a la siguiente pregunta? Si la conoces y estás seguro de ella, responde únicamente "Sí". Si no la conoces o no estás seguro, responde únicamente "No".\\
    Pregunta: \{question\}}
    \end{promptbox}
    \caption{Meta question prompt in Spanish.}
    \label{fig:meta-question-prompt-es}
\end{figure}

\clearpage
\section*{NeurIPS Paper Checklist}


\begin{enumerate}

\item {\bf Claims}
    \item[] Question: Do the main claims made in the abstract and introduction accurately reflect the paper's contributions and scope?
    \item[] Answer: \answerYes{} 
    \item[] Justification: The abstract and introduction clearly state the claims regarding the proposed metacognitive measurement framework ($d'_{\rm type2}$) and the Evolution Strategy for Metacognitive Alignment (ESMA) method. These claims match the experimental results demonstrating improved metacognitive sensitivity across various models in \cref{sec:experiments}.

\item {\bf Limitations}
    \item[] Question: Does the paper discuss the limitations of the work performed by the authors?
    \item[] Answer: \answerYes{} 
    \item[] Justification: The paper discusses the limitations of the work in \cref{sec:discussion}. Specifically, the authors acknowledge that despite the success of the ESMA training method, the absolute metacognition level of the fine-tuned models still remains far below that of humans, particularly given the relatively low metacognitive demands of the evaluated QA tasks.

\item {\bf Theory assumptions and proofs}
    \item[] Question: For each theoretical result, does the paper provide the full set of assumptions and a complete (and correct) proof?
    \item[] Answer: \answerNA{} 
    \item[] Justification: The paper is primarily empirical and methodological, focusing on evaluating and fine-tuning the metacognitive capabilities of LLMs rather than proposing formal theoretical proofs.

    \item {\bf Experimental result reproducibility}
    \item[] Question: Does the paper fully disclose all the information needed to reproduce the main experimental results of the paper to the extent that it affects the main claims and/or conclusions of the paper (regardless of whether the code and data are provided or not)?
    \item[] Answer: \answerYes{} 
    \item[] Justification: The paper explicitly details the models used, datasets, evaluation metrics, and experimental setups in \cref{sec:experiments}. Furthermore, exact prompts and hyperparameter settings are provided in \cref{sec:experimental-details}.
    \item[] Guidelines:
    \begin{itemize}
        \item The answer \answerNA{} means that the paper does not include experiments.
        \item If the paper includes experiments, a \answerNo{} answer to this question will not be perceived well by the reviewers: Making the paper reproducible is important, regardless of whether the code and data are provided or not.
        \item If the contribution is a dataset and\slash or model, the authors should describe the steps taken to make their results reproducible or verifiable. 
        \item Depending on the contribution, reproducibility can be accomplished in various ways. For example, if the contribution is a novel architecture, describing the architecture fully might suffice, or if the contribution is a specific model and empirical evaluation, it may be necessary to either make it possible for others to replicate the model with the same dataset, or provide access to the model. In general. releasing code and data is often one good way to accomplish this, but reproducibility can also be provided via detailed instructions for how to replicate the results, access to a hosted model (e.g., in the case of a large language model), releasing of a model checkpoint, or other means that are appropriate to the research performed.
        \item While NeurIPS does not require releasing code, the conference does require all submissions to provide some reasonable avenue for reproducibility, which may depend on the nature of the contribution. For example
        \begin{enumerate}
            \item If the contribution is primarily a new algorithm, the paper should make it clear how to reproduce that algorithm.
            \item If the contribution is primarily a new model architecture, the paper should describe the architecture clearly and fully.
            \item If the contribution is a new model (e.g., a large language model), then there should either be a way to access this model for reproducing the results or a way to reproduce the model (e.g., with an open-source dataset or instructions for how to construct the dataset).
            \item We recognize that reproducibility may be tricky in some cases, in which case authors are welcome to describe the particular way they provide for reproducibility. In the case of proprietary models, it may be that access to the model is limited in some way (e.g., to registered users), but it should be possible for other researchers to have some path to reproducing or verifying the results.
        \end{enumerate}
    \end{itemize}

\item {\bf Open access to data and code}
    \item[] Question: Does the paper provide open access to the data and code, with sufficient instructions to faithfully reproduce the main experimental results, as described in supplemental material?
    \item[] Answer: \answerYes{}{} 
    \item[] Justification: The paper states in \cref{subsec:hyperparameters} that the source code used for the experiments is attached as supplementary material and publicly available at our Github project page.

\item {\bf Experimental setting/details}
    \item[] Question: Does the paper specify all the training and test details (e.g., data splits, hyperparameters, how they were chosen, type of optimizer) necessary to understand the results?
    \item[] Answer: \answerYes{} 
    \item[] Justification: Extensive details on the training dynamics, data splits, and hyperparameters (such as mutation strength, learning rate, population size, and number of iterations) are specified in \cref{subsec:hyperparameters}.

\item {\bf Experiment statistical significance}
    \item[] Question: Does the paper report error bars suitably and correctly defined or other appropriate information about the statistical significance of the experiments?
    \item[] Answer: \answerYes{} 
    \item[] Justification: The paper reports statistical significance testing when comparing ESMA against baseline methods (SFT, PPO, GRPO), using bootstrapping with 10,000 iterations and providing 99\% confidence intervals in \cref{subsec:sft}.

\item {\bf Experiments compute resources}
    \item[] Question: For each experiment, does the paper provide sufficient information on the computer resources (type of compute workers, memory, time of execution) needed to reproduce the experiments?
    \item[] Answer: \answerYes{} 
    \item[] Justification: \cref{subsec:hyperparameters} provides sufficient information on the compute resources, explicitly noting the use of eight NVIDIA Quadro RTX 6000 24GB GPUs or one NVIDIA A100 80GB GPU, alongside batch sizes and training steps for the various methods.
    
\item {\bf Code of ethics}
    \item[] Question: Does the research conducted in the paper conform, in every respect, with the NeurIPS Code of Ethics \url{https://neurips.cc/public/EthicsGuidelines}?
    \item[] Answer: \answerYes{} 
    \item[] Justification: The research adheres to standard ethical guidelines for machine learning research without any specialized deviations.

\item {\bf Broader impacts}
    \item[] Question: Does the paper discuss both potential positive societal impacts and negative societal impacts of the work performed?
    \item[] Answer: \answerYes{} 
    \item[] Justification: A `Broader impacts' section is included in \cref{sec:impact-statement}, which discusses the positive societal benefits of reducing AI misinformation while also explicitly addressing potential dual-use risks, such as the misuse of enhanced metacognitive awareness for sophisticated deception.
    
\item {\bf Safeguards}
    \item[] Question: Does the paper describe safeguards that have been put in place for responsible release of data or models that have a high risk for misuse (e.g., pre-trained language models, image generators, or scraped datasets)?
    \item[] Answer: \answerNA{} 
    \item[] Justification: The paper involves fine-tuning existing open-source models for improved metacognition and does not release new, high-risk foundational models or newly scraped datasets requiring specific safety filters.

\item {\bf Licenses for existing assets}
    \item[] Question: Are the creators or original owners of assets (e.g., code, data, models), used in the paper, properly credited and are the license and terms of use explicitly mentioned and properly respected?
    \item[] Answer: \answerYes{} 
    \item[] Justification: All existing assets, including the base LLMs (Qwen2.5, Llama3.2, Gemma3) and datasets (TriviaQA, FreebaseQA, etc.), are appropriately credited and cited throughout the paper and in the References section.

\item {\bf New assets}
    \item[] Question: Are new assets introduced in the paper well documented and is the documentation provided alongside the assets?
    \item[] Answer: \answerYes{} 
    \item[] Justification: The code, methodologies (ESMA), and specific prompt templates introduced in this research are documented in the main text and \cref{subsec:hyperparameters}, and are included in the supplementary material.

\item {\bf Crowdsourcing and research with human subjects}
    \item[] Question: For crowdsourcing experiments and research with human subjects, does the paper include the full text of instructions given to participants and screenshots, if applicable, as well as details about compensation (if any)? 
    \item[] Answer: \answerNA{} 
    \item[] Justification: The study does not involve human subjects or crowdsourcing.

\item {\bf Institutional review board (IRB) approvals or equivalent for research with human subjects}
    \item[] Question: Does the paper describe potential risks incurred by study participants, whether such risks were disclosed to the subjects, and whether Institutional Review Board (IRB) approvals (or an equivalent approval/review based on the requirements of your country or institution) were obtained?
    \item[] Answer: \answerNA{} 
    \item[] Justification: The research does not involve human subjects.

\item {\bf Declaration of LLM usage}
    \item[] Question: Does the paper describe the usage of LLMs if it is an important, original, or non-standard component of the core methods in this research? Note that if the LLM is used only for writing, editing, or formatting purposes and does \emph{not} impact the core methodology, scientific rigor, or originality of the research, declaration is not required.
    \item[] Answer: \answerNA{} 
    \item[] Justification: The core method development in this research does not involve LLMs as an active tool or non-standard component. Any usage of LLMs in the preparation of this manuscript was limited to writing, editing, and routine coding assistance purposes, which does not require declaration under the NeurIPS policy.

\end{enumerate}

\end{document}